\documentclass[lettersize,journal]{IEEEtran}
\usepackage{amsmath,amsfonts}
\usepackage{array}
\usepackage[caption=false,font=normalsize,labelfont=sf,textfont=sf]{subfig}
\usepackage{textcomp}
\usepackage{stfloats}
\usepackage{url}
\usepackage{verbatim}
\usepackage{graphicx}
\usepackage{cite}

\usepackage{booktabs}       % professional-quality tables
\usepackage{nicefrac}       % compact symbols for 1/2, etc.
\usepackage{microtype}      % microtypography
\usepackage{xcolor}         % colors
\usepackage{xspace}
\usepackage{wrapfig}
\usepackage{amssymb}
\usepackage{bm}
\usepackage{tabularx}
\usepackage{paralist}
\usepackage{mathtools}
\usepackage[linesnumbered,ruled,vlined]{algorithm2e}
\usepackage{pifont}
\usepackage{multirow}

\makeatletter
\DeclareRobustCommand\onedot{\futurelet\@let@token\@onedot}
\def\@onedot{\ifx\@let@token.\else.\null\fi\xspace}
\def\eg{\emph{e.g}\onedot} 
\def\ie{\emph{i.e}\onedot}

\def\etal{\emph{et al}\onedot}
\makeatother

\newcommand{\Tref}[1]{Table~\ref{#1}}
\newcommand{\Eref}[1]{Equation~(\ref{#1})}
\newcommand{\Fref}[1]{Figure~\ref{#1}}

\begin{document}

\title{High-Speed Full-Color HDR Imaging via Unwrapping Modulo-Encoded Spike Streams}

\author{
Chu Zhou$^\#$, Siqi Yang$^\#$, Kailong Zhang, Heng Guo, Zhaofei Yu,~\IEEEmembership{Member,~IEEE}, Boxin Shi,~\IEEEmembership{Senior Member,~IEEE} and Imari Sato,~\IEEEmembership{Member,~IEEE}
    \thanks{
    $^\#$ Equal contribution.
    }
    \thanks{
    Chu Zhou and Imari Sato are with the Digital Content and Media Sciences Research Division, National Institute of Informatics, Tokyo 101-8430, Japan.
    }
    \thanks{
    Kailong Zhang and Heng Guo are with the Pattern Recognition and Intelligent System Laboratory, School of Artificial Intelligence, Beijing University of Posts and Telecommunications, Beijing 100876, China
    }
    \thanks{
    Boxin Shi is with the State Key Laboratory of Multimedia Information Processing, School of Computer Science, the National Engineering Research Center of Visual Technology, School of Computer Science, and the PKU-AI$^2$ Robotics Joint Lab of Embodied AI, Peking University, Beijing 100080, China. 

    Siqi Yang is with the Institute for Artificial Intelligence, the State Key Laboratory of Multimedia Information Processing, School of Computer Science, and the National Engineering Research Center of Visual Technology, School of Computer Science, Peking University, Beijing 100080, China.
    
    Zhaofei Yu is with the Institute for Artificial Intelligence, and the National Engineering Research Center of Visual Technology, School of Computer Science, Peking University, Beijing 100080, China. 
    }
}

\markboth{IEEE TRANSACTIONS ON PATTERN ANALYSIS AND MACHINE INTELLIGENCE}%
{Shell \MakeLowercase{\textit{et al.}}: A Sample Article Using IEEEtran.cls for IEEE Journals}

\IEEEpubid{0000--0000/00\$00.00~\copyright~2021 IEEE}

\maketitle

\begin{abstract}
Conventional RGB-based high dynamic range (HDR) imaging faces a fundamental trade-off between motion artifacts in multi-exposure captures and irreversible information loss in single-shot techniques. Modulo sensors offer a promising alternative by encoding theoretically unbounded dynamic range into wrapped measurements. However, existing modulo solutions remain bottlenecked by iterative unwrapping overhead and hardware constraints limiting them to low-speed, grayscale capture. In this work, we present a complete modulo-based HDR imaging system that enables high-speed, full-color HDR acquisition by synergistically advancing both the sensing formulation and the unwrapping algorithm. At the core of our approach is an exposure-decoupled formulation of modulo imaging that allows multiple measurements to be interleaved in time, preserving a clean, observation-wise measurement model. Building upon this, we introduce an iteration-free unwrapping algorithm that integrates diffusion-based generative priors with the physical least absolute remainder property of modulo images, supporting highly efficient, physics-consistent HDR reconstruction. Finally, to validate the practical viability of our system, we demonstrate a proof-of-concept hardware implementation based on modulo-encoded spike streams. This setup preserves the native high temporal resolution of spike cameras, achieving 1000 FPS full-color imaging while reducing output data bandwidth from approximately 20 Gbps to 6 Gbps. Extensive evaluations indicate that our coordinated approach successfully overcomes key systemic bottlenecks, demonstrating the feasibility of deploying modulo imaging in dynamic scenarios.
\end{abstract}


\begin{figure*}[t]
    \centering
    \includegraphics[width=1.0\linewidth]{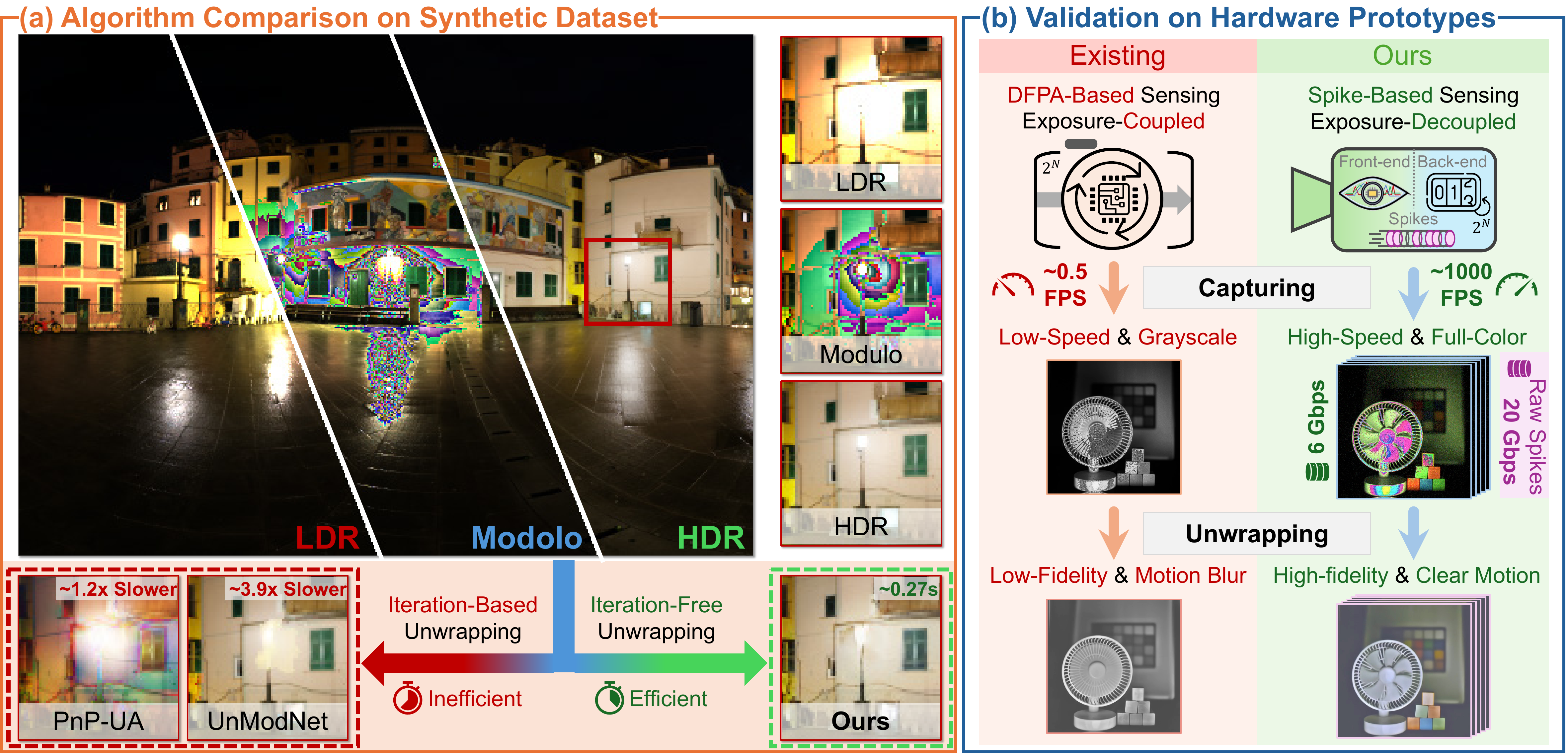}
    \caption{We present a complete modulo-based HDR imaging system capable of high-speed, full-color HDR acquisition. (a) Algorithm comparison on the synthetic dataset \cite{zhou2020unmodnet}: Compared to state-of-the-art iteration-based unwrapping approaches (UnModNet \cite{zhou2020unmodnet} and PnP-UA \cite{bacca2024deep}), our iteration-free unwrapping algorithm achieves both superior visual quality and the fastest reconstruction speed ($\sim0.27$s per frame). (b) Validation on hardware prototypes: Existing prototypes \cite{zhao2015unbounded} rely on DFPA-based sensing and exposure-coupled formulations, which are limited to low-speed ($\sim0.5$ FPS), grayscale captures prone to motion blur. In contrast, our proof-of-concept prototype introduces a spike-based sensing scheme with an exposure-decoupled formulation. This enables high-speed ($\sim1000$ FPS), full-color modulo imaging with high-fidelity motion clarity, while significantly reducing the transmission bandwidth from 20 Gbps (raw spike baseline \cite{yang2024real}) to 6 Gbps.}
    \label{fig: Teaser}
\end{figure*}

\section{Introduction}
\IEEEPARstart{R}{eal}-world environments frequently exhibit extreme illumination variations, where the dynamic range of a single scene can easily span several orders of magnitude. Conventional digital cameras, however, possess a strictly limited dynamic range. Consequently, when capturing such high-contrast scenes, standard sensors inevitably suffer from saturation in bright areas and severe noise in dark regions, yielding low dynamic range (LDR) images with an irreversible loss of physical irradiance information. To accurately record the full extent of scene radiance, high dynamic range (HDR) imaging has become an essential area of study, vital not only for preserving perceptual visual quality but also for ensuring the robustness of downstream computer vision tasks.

A common paradigm to achieve HDR imaging is to computationally reconstruct HDR content from conventional LDR observations. The most prevalent approach in this vein is multi-exposure image fusion \cite{debevec1997recovering}, which merges a sequence of LDR images captured under bracketed exposure times. This multi-shot strategy is highly effective for recovering rich details and expanding the dynamic range in static environments. However, it is inherently fragile in the presence of camera motion or scene dynamics. Despite extensive efforts to develop misalignment-aware algorithms \cite{sen2012robust, kalantari2017deep}, resolving complex occlusions and mitigating severe ghosting artifacts remains a persistent challenge. Alternatively, to circumvent the issues associated with sequential acquisition, another line of research focuses on single-image HDR reconstruction \cite{eilertsen2017hdr, endo2017deep, santos2020single}. While this single-shot approach successfully avoids motion artifacts entirely, hallucinating lost content from a single clipped LDR observation is a severely ill-posed inverse problem. Without deterministic physical measurements in the saturated regions, these algorithms inevitably struggle to recover physically faithful details, fundamentally limiting the achievable reconstruction fidelity and generalizability.

\IEEEpubidadjcol

To bypass these intrinsic bottlenecks of conventional cameras, unconventional sensors have emerged to address the HDR challenge at the hardware level. While coded exposure techniques \cite{nayar2000high, serrano2016convolutional} enable single-shot capture, their optical masks inevitably block incident light, reducing the signal-to-noise ratio. Alternatively, neuromorphic sensors provide a compelling pathway by encoding visual signals as events or spikes, naturally avoiding saturation. Event cameras \cite{brandli2014240} record only logarithmic intensity changes, lacking the absolute irradiance information necessary for standalone high-fidelity HDR reconstruction. Spike cameras \cite{zhu2019retina}, in contrast, emit binary spikes whose temporal density directly reflects scene irradiance, enabling direct HDR imaging. However, this representational advantage incurs a severe system-level cost: the resulting high-rate spike streams are extremely bandwidth-intensive. A typical spike camera operating at $f \geq 20$ kHz requires over $20$ Gbps of transmission bandwidth, precluding practical scalability.

Modulo imaging offers a bandwidth-efficient paradigm for HDR acquisition by periodically wrapping the sensor output instead of allowing it to saturate \cite{zhao2015unbounded}. With appropriate phase unwrapping \cite{zhou2020unmodnet, jagatap2020high, bacca2024deep, chen2025robust}, these measurements can theoretically achieve an unbounded dynamic range. Despite this potential, existing modulo-based systems remain ill-suited for practical high-speed, full-color capture. Current designs are hindered by an \textit{exposure-coupled formulation}: each modulo image requires the completion of a full exposure period, intrinsically capping the achievable frame rate. Furthermore, the corresponding reconstruction processes typically rely on computationally heavy iterative unwrapping, and current hardware prototypes are largely restricted to low-speed, grayscale-only implementations.

In this work, we present a complete modulo-based HDR imaging system that overcomes these compounded bottlenecks, enabling high-speed, full-color HDR acquisition. At the core of our approach is an \textit{exposure-decoupled formulation} of modulo imaging that conceptually decomposes the sensing process into two independent phases: representation and query. Rather than tying each modulo image to a full exposure, our formulation generates observations by querying a temporally dense signal representation, allowing multiple modulo measurements to be interleaved in time. This decoupling breaks the exposure-imposed limit, enabling high-speed capture (up to $1000$ FPS) while preserving a clean, observation-wise forward model. As shown in \Fref{fig: Teaser} (a), on the algorithmic side, we propose an iteration-free modulo unwrapping framework that leverages the strong generative priors of a pre-trained diffusion model while explicitly exploiting the least absolute remainder property of modulo images. This ensures a strict physical consistency constraint, enabling efficient, per-observation HDR reconstruction with an inference time of approximately $0.27$s per frame. As shown in \Fref{fig: Teaser} (b), on the hardware side, rather than fabricating a custom silicon sensor, we physically reconfigure an off-the-shelf spike camera to instantiate our exposure-decoupled formulation. The resulting architecture consists of a high-speed chromatic spike sensing front-end with a non-Bayer sampling pattern, coupled with a per-pixel modulo encoding back-end based on parallel register arrays. By shifting the transmission bottleneck from raw spike streaming to modulo-based encoding, our setup drastically compresses the output data rate from approximately $20$ Gbps ($2.5$ GB/s) to $6$ Gbps ($0.75$ GB/s).

In summary, this paper makes the following contributions:
\begin{itemize}
    \item An \textbf{exposure-decoupled formulation} of modulo imaging, which breaks traditional frame-rate barriers to ensure high-speed capture while preserving per-observation fidelity.
    \item An \textbf{iteration-free algorithm} for modulo unwrapping, which integrates generative priors with deterministic physical constraints to support highly efficient and physics-consistent HDR reconstruction.
    \item A \textbf{bandwidth-efficient hardware} implementation, which reconfigures high-speed spike sensing into a modulo paradigm, successfully reducing data rates by over $70\%$ and enabling full-color HDR acquisition.
\end{itemize}

\section{Related Work}
High dynamic range (HDR) imaging has been extensively studied over the past decades. Existing approaches can be broadly categorized into two classes based on the imaging modality: methods built upon conventional RGB image sensors and those relying on unconventional hardware mechanisms. We review representative works in each category below.

\subsection{HDR Imaging from Conventional RGB Cameras}
The most common paradigm for HDR imaging involves capturing LDR images using conventional RGB cameras, followed by algorithmic post-processing to reconstruct the full dynamic range. These approaches are generally divided into multi-image and single-image frameworks.

\textbf{HDR imaging from multiple LDR images.} The classic approach constructs an HDR image by merging multiple LDR captures taken under bracketed exposures \cite{debevec1997recovering}. While effective in static scenes, camera motion or scene dynamics frequently introduce spatial misalignment, leading to severe ghosting artifacts. Early methods mitigated this via misalignment-aware numerical optimization \cite{khan2006ghost, sen2012robust, oh2014robust}. More recently, learning-based frameworks have significantly improved reconstruction quality by jointly addressing exposure alignment and fusion \cite{kalantari2017deep, prabhakar2020towards}. To better handle large motions and preserve fine details, subsequent works incorporated adversarial learning \cite{niu2021hdr} and attention mechanisms \cite{yan2022dual}. Advanced learning paradigms, including diffusion models \cite{hu2024generating}, guided inpainting \cite{chen2025ultrafusion}, and deep unfolding frameworks \cite{li2025afunet}, have further pushed the limits of visual fidelity. Related extensions have also explored mobile photography under severe noise \cite{hasinoff2016burst, liu2023joint} and HDR video reconstruction \cite{kalantari2019deep, xu2024hdrflow}. Despite these advances, multi-exposure methods inherently increase capture complexity and remain fundamentally susceptible to high-speed motion.

\textbf{HDR imaging from a single LDR image.} To entirely avoid ghosting artifacts, another line of work—often referred to as inverse tone mapping—aims to recover HDR content from a single LDR observation \cite{banterle2006inverse}. Early approaches relied on numerical optimization and handcrafted priors derived from natural image statistics \cite{banterle2006inverse, masia2009evaluation, rempel2007ldr2hdr}. With the advent of deep learning, modern methods directly infer missing HDR details by predicting virtual bracketed exposures \cite{endo2017deep}, end-to-end feature-level fusing \cite{eilertsen2017hdr, marnerides2018expandnet, santos2020single}, or utilizing physically motivated formulations that invert the HDR-to-LDR pipeline \cite{liu2020single}. Recent pipelines have also leveraged intrinsic image decomposition \cite{dille2024intrinsic} and generative diffusion models \cite{wang2025lediff} to hallucinate lost details. However, this single-shot paradigm faces a severely ill-posed reconstruction problem; the irreversible loss of information in saturated or under-exposed regions fundamentally bottlenecks the achievable reconstruction fidelity.

\subsection{HDR Imaging via Unconventional Sensors} 
To overcome the limitations of conventional RGB sensors, unconventional imaging hardware has been developed to enable single-shot, ghost-free HDR reconstruction with high fidelity. We categorize these approaches into three main groups based on their underlying principles.

\textbf{HDR imaging from coded exposure.} Coded exposure systems augment conventional image sensors with optical masks to induce spatially varying exposures within a single shot. Early research primarily focused on static optical mask design—using regular \cite{nayar2000high} or non-regular \cite{schoberl2012high} attenuation patterns—and classical reconstruction algorithms like piecewise linear estimation \cite{aguerrebere2014single} or convolutional sparse coding \cite{serrano2016convolutional}. Recent studies have transitioned toward reconfigurable coded exposure systems \cite{alghamdi2019reconfigurable} and end-to-end optical-electronic co-design, where hardware encoding and neural reconstruction are jointly optimized \cite{metzler2020deep}. Similar spatially varying exposure concepts have also been realized using division-of-focal-plane micro-polarizers in polarization sensors \cite{zhou2023polarizationHDR}. Nevertheless, these approaches share a common physical drawback: the optical masks inevitably attenuate incoming light, resulting in a reduced signal-to-noise ratio.

\textbf{HDR imaging from neuromorphic sensors.} Neuromorphic sensors, including event cameras \cite{brandli2014240} and spike cameras \cite{zhu2019retina}, encode visual signals into high-rate event or spike streams instead of capturing conventional intensity frames. This mechanism naturally avoids saturation and enables extreme dynamic range capture. Prior work has exploited these representations either by directly reconstructing HDR intensity frames from neuromorphic data \cite{rebecq2019high, wang2021asynchronous, yang2024real}, or by utilizing these neuromorphic data to guide the HDR reconstruction of standard frame-based LDR images \cite{han2020neuromorphic, han2023hybrid, chang20231000}. While neuromorphic approaches are highly promising, they involve inherent trade-offs: event-based methods benefit from low data rates but lack absolute irradiance measurements—limiting standalone reconstruction fidelity—whereas spike-based methods enable better direct frame reconstruction but incur excessively high data bandwidths, restricting scalability.

\textbf{HDR imaging from modulo sensors.} Modulo sensors offer a theoretically unbounded dynamic range by periodically wrapping the accumulated photocharge whenever a predefined saturation threshold is reached during exposure \cite{zhao2015unbounded}. Early methods tackled the resulting phase unwrapping problem using numerical optimization \cite{zhao2015unbounded, jagatap2020high}, demonstrating the potential for extreme HDR capture. To improve robustness, subsequent methods introduced learning-based unwrapping, employing direct neural network mapping \cite{zhou2020unmodnet}, deep plug-and-play priors \cite{bacca2024deep}, and deep unfolding strategies \cite{chen2025robust}. Despite these advances, existing modulo-based HDR frameworks remain bottlenecked by the gap between theory and physical capture. Systemically, many rely on exposure-coupled formulations and are evaluated exclusively on simulated data. Meanwhile, the few existing proof-of-concept prototypes are restricted to low-speed, grayscale capture. Furthermore, their reconstruction algorithms typically suffer from severe iterative unwrapping overhead. These limitations restrict practical deployment, highlighting the need for a high-speed, full-color HDR imaging paradigm that synergistically coordinates efficient unwrapping with practical prototype configurations.

\section{Exposure-Decoupled Formulation of Modulo Imaging}
\subsection{Modulo Imaging: Background} 
The fundamental principle of modulo imaging is to periodically wrap the accumulated photocharge (or equivalent digital signal) upon reaching a predefined saturation threshold, thereby preserving high-fidelity information in otherwise over-exposed regions. Let $\mathbf{I}$ denote the exposure-integrated irradiance mapped to digital counts by an idealized linear imaging pipeline with an unlimited dynamic range, which corresponds to the ideal HDR image we aim to reconstruct. Following the pioneering formulation of modulo imaging by Zhao \etal \cite{zhao2015unbounded}, we assume that the exposure duration is pre-calibrated such that low-irradiance regions are properly exposed. By explicitly focusing on resolving unbounded dynamic range at the bright end, this assumption ensures a reliable signal-to-noise ratio in darker areas and abstracts away the low-light denoising problem.

In a standard modulo sensor, instead of clipping the integrated signal once it reaches the maximum representable level, the digital value is wrapped modulo $2^N$, where $N$ denotes the bit depth of the sensor. The recorded modulo image is thus mathematically given by:
\begin{equation}
    \mathbf{I}_m = \text{mod}(\mathbf{I}, 2^N).
\end{equation}
By retaining the wrapped measurements in excessively bright regions while leaving the properly exposed low-irradiance regions unaltered, modulo imaging captures a drastically expanded dynamic range within a single observation. Once $\mathbf{I}_m$ is available, an unwrapping algorithm can be employed to recover the ideal $\mathbf{I}$. Because the wrapped values provide deterministic physical constraints, this unwrapping process is generally much better-posed than attempting to hallucinate irreversibly lost details from conventionally clipped LDR images.

\begin{figure}[t]
  \centering
  \includegraphics[width=1.0\linewidth]{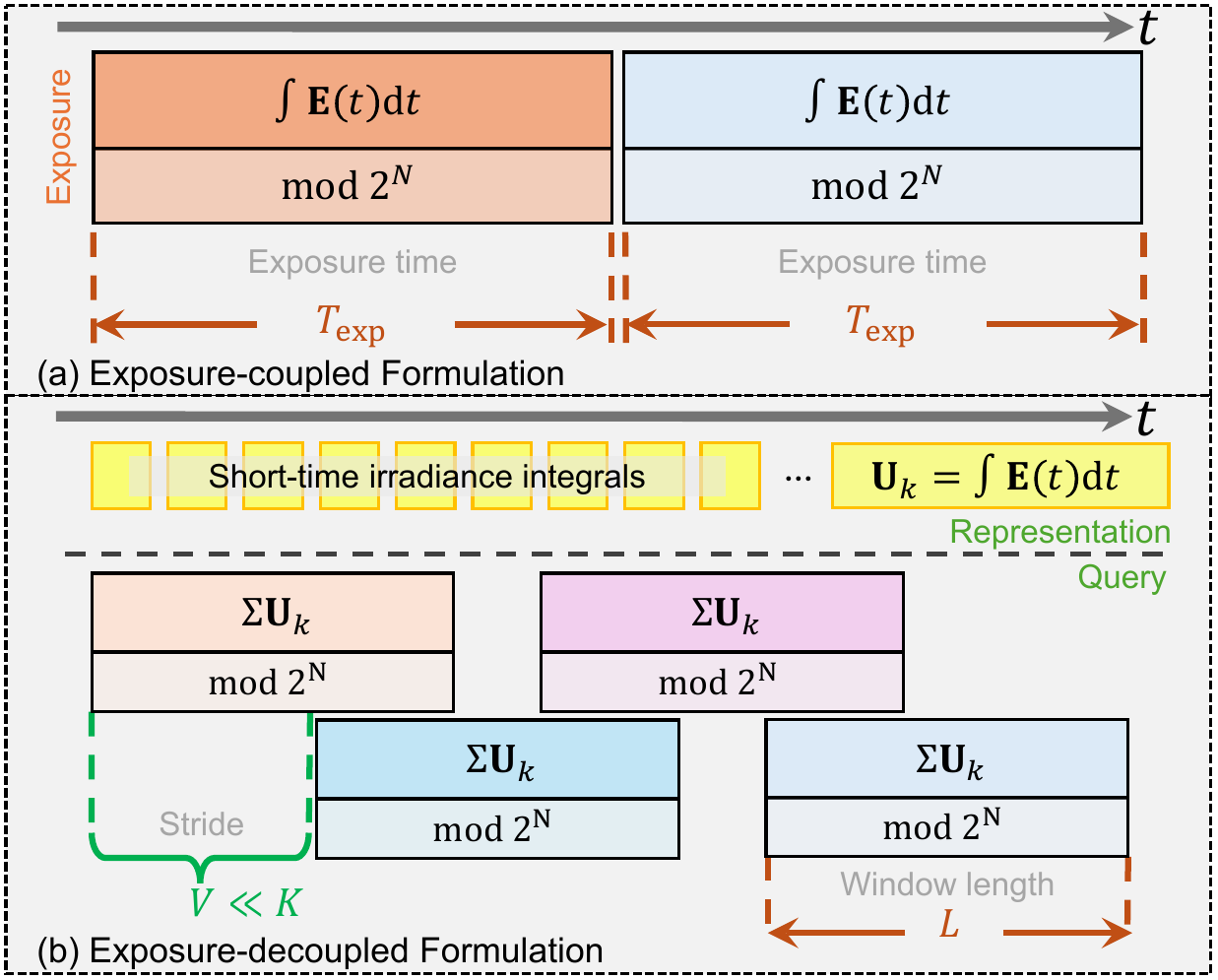}
  \caption{Illustration of the difference between modulo imaging formulations. (a) In existing exposure-coupled formulations, each modulo image is generated by completing a full exposure over a fixed temporal interval. (b) In our exposure-decoupled formulation, the imaging process is decomposed into two independent phases, namely representation and query, which explicitly decouples temporal signal integration from modulo image generation.}
  \label{fig: Formulation}
\end{figure}

\textbf{Existing exposure-coupled formulation.} In existing modulo-based HDR imaging systems, each modulo image is strictly generated by completing an entire physical exposure over a fixed temporal interval, as shown in \Fref{fig: Formulation} (a). Let $\mathbf{E}(t)$ denote the continuous incident scene irradiance. The $i$-th modulo image, starting at time $t_i$, is expressed as:
\begin{equation}
    \mathbf{I}_m^{(i)} = \text{mod} \left( \left\lfloor h \int_{t_i}^{t_{i+1}} \mathbf{E}(t) \mathrm{d}t \right\rfloor, 2^N \right), \; t_{i+1} = t_i + T_\text{exp},
\end{equation}
where $T_\text{exp}$ denotes the physical exposure duration and $h$ represents the radiometric-to-digital conversion gain. Under this conventional paradigm, the achievable capture frame rate is intrinsically bounded by $1/T_\text{exp}$. If one attempts to capture high-speed motion by reducing $T_\text{exp}$, the total integrated signal drops proportionally. This severely degrades the fidelity in low-irradiance regions, directly violating the proper-exposure assumption mentioned above. Consequently, this tight coupling between temporal signal integration and modulo image generation imposes a fundamental and restrictive trade-off between the temporal resolution (frame rate) and the radiometric signal quality.

\subsection{Exposure-Decoupled Formulation}
To completely overcome the frame-rate limitations imposed by the aforementioned exposure coupling, we propose to explicitly decouple the temporal signal integration from the generation of modulo images. As illustrated in \Fref{fig: Formulation} (b), we achieve this by conceptually decomposing the sensing process into two independent phases: \emph{representation} and \emph{query}.

In the \emph{representation} phase, we assume access to a temporally dense, unquantized integral representation of the scene dynamics. Specifically, the total capturing time $T$ is uniformly discretized into $K$ extremely short micro-intervals. This yields a temporally dense sequence $\{\mathbf{U}_k\}_{k=1}^{K}$, where the $k$-th element denotes the irradiance integrated strictly over the $k$-th interval:
\begin{equation}
    \mathbf{U}_k = \int_{\tau_k}^{\tau_k + T / K} \mathbf{E}(\tau) \mathrm{d}\tau, \; \tau_k=(k-1)\frac{T}{K}.
\end{equation}
Here, the parameter $K$ governs the underlying temporal resolution of the system and can be chosen to be sufficiently large to capture highly transient dynamics, without imposing any artificial constraints on the effective exposure duration of the final modulo measurements.

In the subsequent \emph{query} phase, actual modulo observations are generated by querying this dense representation over multiple, potentially overlapping, temporal sliding windows. Each window spans $L$ consecutive sub-intervals and advances with a fixed temporal stride $V$. For each index $i$, the starting index of the window is $b_i = (i-1)V + 1$, and the corresponding modulo image is formulated as:
\begin{equation}
    \label{eq: OurFormulation}
    \mathbf{I}_m^{(i)} = \text{mod} \left( \left\lfloor h \sum_{k=b_i}^{b_i + L - 1} \mathbf{U}_k \right\rfloor, 2^N \right).
\end{equation}
Under this decoupled formulation, the effective exposure duration for each individual modulo image remains $LT/K$ (ensuring sufficient light collection), yet the effective output frame rate becomes $K/VT$ instead of the coupled $K/LT$. This mathematically demonstrates that the output frame rate (determined by stride $V$) and the exposure duration (determined by window length $L$) are no longer interdependent. Consequently, $L$ can be kept sufficiently large to guarantee a robust signal-to-noise ratio in low-irradiance regions, while $V$ can be minimized to produce a high-speed sequence.

Crucially, unlike temporally multiplexed or coded sensing schemes that scramble the physical meaning of the measurements, our formulation preserves a clean, observation-wise forward model. Each generated measurement remains directly interpretable as a standard, physically valid modulo image, providing a consistent and mathematically sound foundation for both our subsequent algorithm design and proof-of-concept hardware implementation.

\begin{figure*}[t]
  \centering
  \includegraphics[width=\linewidth]{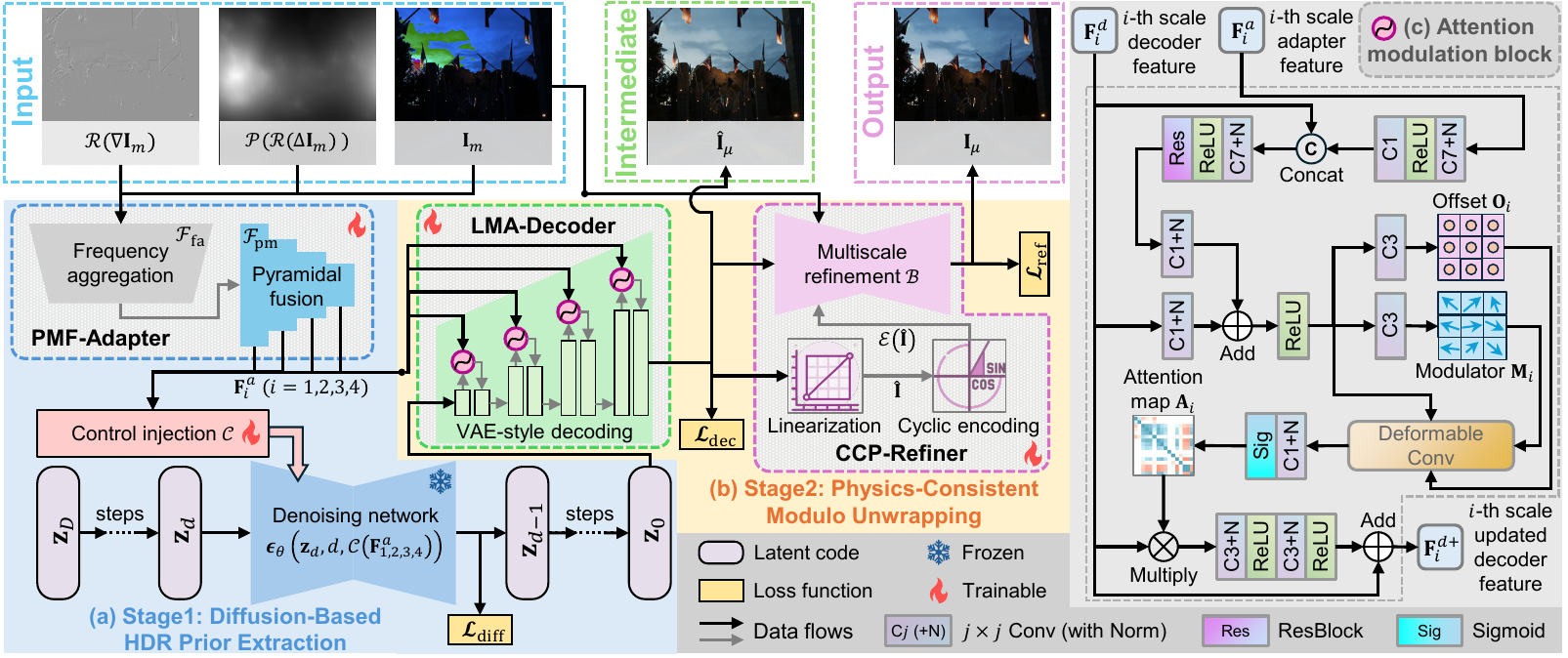}
  \caption{Illustration of the proposed iteration-free modulo unwrapping framework, which consists of two stages: (a) diffusion-based HDR prior extraction and (b) physics-consistent modulo unwrapping. In Stage1, a pyramidal multi-frequency adapter (PMF-Adapter) extracts multiscale features $\mathbf{F}_{1,2,3,4}^a$ from the input modulo image $\mathbf{I}_m$ and its derivative representations, capturing both low- and high-frequency scene information. These features are injected into a pre-trained diffusion model at corresponding scales to condition the denoising process, yielding a latent code $\mathbf{z}_0$ that serves as an HDR prior in the $\mu$-law tone-mapped domain. In Stage2, a latent-modulated attention decoder (LMA-Decoder) reconstructs an initial HDR estimate from $\mathbf{z}_0$ under the guidance of $\mathbf{F}_{1,2,3,4}^a$, followed by a cyclic-consistent physical refiner (CCP-Refiner) that enforces physical consistency constraint and produces the final HDR output. (c) shows the detailed architecture of the proposed attention modulation block (AMB) in the LMA-Decoder.}
  \label{fig: Algorithm}
\end{figure*}

\section{Iteration-Free Unwrapping Algorithm}
Existing modulo-based HDR imaging systems typically rely on iterative unwrapping in the linear intensity domain, which is not only inefficient but also prone to error accumulation. Recent diffusion models \cite{ho2020denoising, song2020denoising} provide a promising alternative by capturing rich natural image statistics, enabling the direct inference of plausible HDR content without explicit wrap counting. However, diffusion-based inference alone may induce hallucination, fundamentally violating the physical constraints of modulo imaging. To solve this issue, we propose a two-stage framework that integrates diffusion-based HDR prior extraction with explicit physical consistency enforcement. As shown in \Fref{fig: Algorithm}, the framework is structured into (a) diffusion-based HDR prior extraction and (b) physics-consistent modulo unwrapping, enabling iteration-free HDR reconstruction while preserving fidelity to the forward model of modulo imaging.

\subsection{Background on Diffusion Models}
For completeness, we briefly review the basics of diffusion models, which form the generative prior underlying our unwrapping algorithm. Diffusion models \cite{ho2020denoising, song2020denoising} learn complex data distributions by reversing a gradual noising process. The forward diffusion process is formulated as a Markov chain that progressively adds Gaussian noise to a clean image. Using $\mathcal{N}$ to denote a Gaussian distribution, starting from $\mathbf{x}_0 \sim p_{\text{data}}(\mathbf{x})$, the process evolves as:
\begin{equation}
    q(\mathbf{x}_d \mid \mathbf{x}_{d-1}) = \mathcal{N} \! \left(\mathbf{x}_d; \sqrt{\alpha_d} \mathbf{x}_{d-1}, (1-\alpha_d) \mathbf{1} \right),
\end{equation}
where $d \in \{1, \dots, D\}$ denotes the diffusion step, and $\{\alpha_d\}_{d=1}^{D}$ is a predefined noise schedule controlling the injected variance. By recursively applying the transition, $\mathbf{x}_d$ admits a closed-form expression with respect to $\mathbf{x}_0$:
\begin{equation}
    \mathbf{x}_d = \sqrt{\bar{\alpha}_d} \mathbf{x}_0 + \sqrt{1-\bar{\alpha}_d} \boldsymbol{\epsilon},
\end{equation}
where $\boldsymbol{\epsilon} \sim \mathcal{N}(\mathbf{0}, \mathbf{1})$ denotes the injected noise and $\bar{\alpha}_d = \prod_{i=1}^{d}\alpha_i$. To approximate the intractable reverse denoising process, a neural network $\boldsymbol{\epsilon}_\theta$ (with parameters $\theta$) is trained to predict the injected noise $\boldsymbol{\epsilon}$ from the noisy observation $\mathbf{x}_d$ and step $d$. To improve computational efficiency, latent diffusion models (LDMs) \cite{rombach2022high} perform this process in a compact latent space, where a variational autoencoder (VAE) maps the image $\mathbf{x}_0$ to a latent code $\mathbf{z}_0$. Once trained, $\boldsymbol{\epsilon}_\theta$ captures rich natural image statistics and can be leveraged as a powerful prior for our highly ill-posed unwrapping task.

\subsection{Stage1: Diffusion-Based HDR Prior Extraction} 
As shown in \Fref{fig: Algorithm} (a), this stage aims to extract a reliable HDR prior in the latent space by conditioning a pre-trained diffusion model on the modulo observation. While a ControlNet-style mechanism \cite{zhang2023adding} is conventionally used to inject control features into the denoising network $\boldsymbol{\epsilon}_\theta$, directly using a modulo image as the control signal remains challenging due to the severe nonlinearity introduced by the modulo operation.

\begin{figure}[t]
  \centering
  \includegraphics[width=\linewidth]{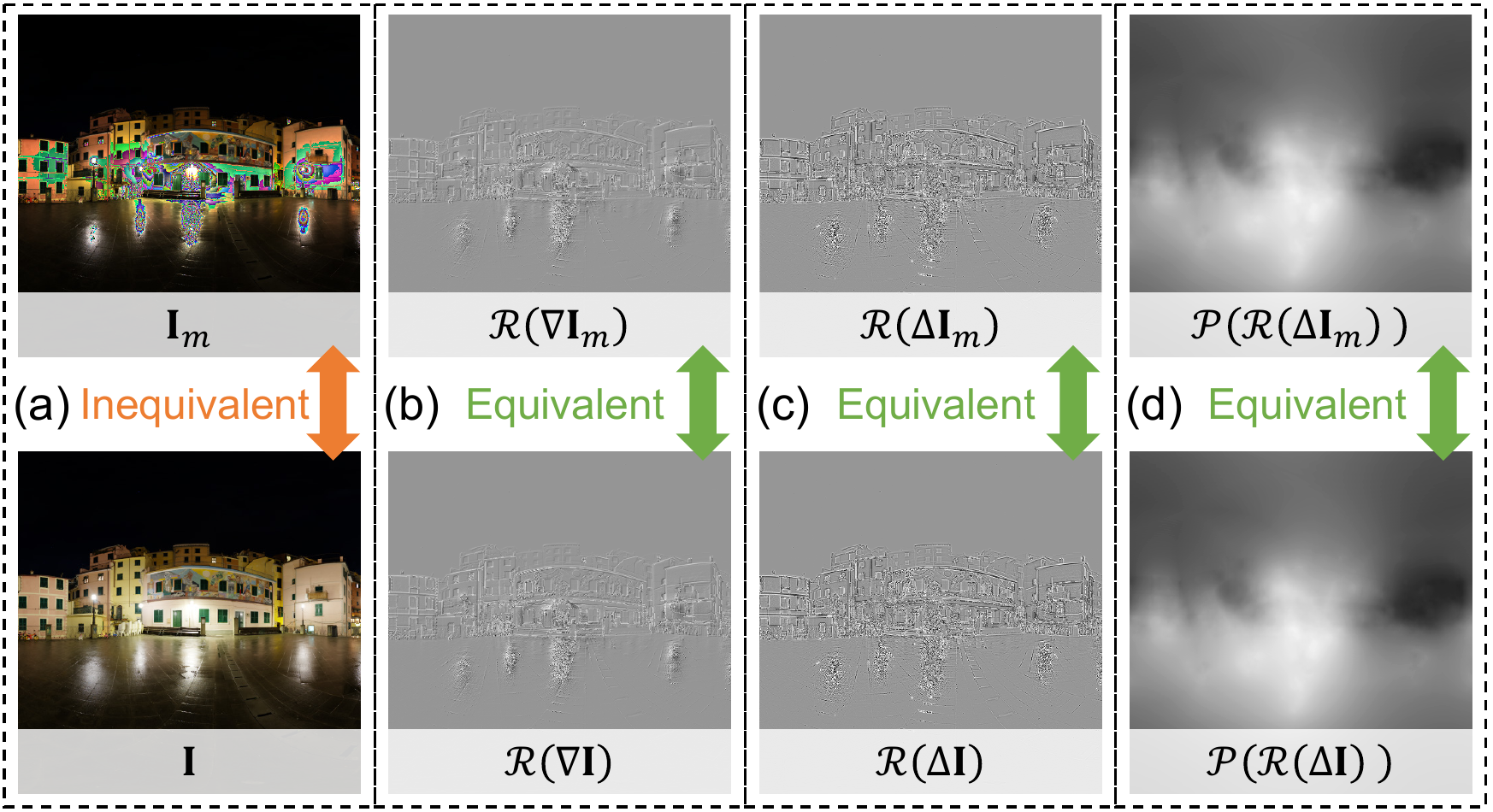}
  \caption{Illustration of the least absolute remainder (LAR) property of modulo images. (a) The modulo image $\mathbf{I}_m$ and its HDR counterpart $\mathbf{I}$ are not equivalent in the intensity domain. (b) After applying the LAR operator $\mathcal{R}$, the two become equivalent in the gradient domain ($\nabla$). (c) This equivalence extends to the Laplacian domain ($\Delta$). (d) The equivalence remains valid after applying a Poisson solver $\mathcal{P}$ to the Laplacian.}
  \label{fig: LAR}
\end{figure}

\textbf{The least absolute remainder (LAR) property.} Prominently, we observe an important structural property of modulo images. As shown in \Fref{fig: LAR} (a) and (b), the modulo image $\mathbf{I}_m$ and the underlying HDR image $\mathbf{I}$ become equivalent in the gradient domain after applying the least absolute remainder (LAR) operation $\mathcal{R}$:
\begin{equation}
    \mathcal{R}(\nabla \mathbf{I}_m) = \mathcal{R}(\nabla \mathbf{I}),
\end{equation}
where $\nabla$ is the gradient operator. The LAR operator $\mathcal{R}$ is defined as:
\begin{equation}
    \mathcal{R}(\mathbf{o}) = \text{mod} \left(\mathbf{o} + \frac{m}{2}, m\right) - \frac{m}{2}, \; m = 2^N.
\end{equation}
This property can be viewed as a generalized form of Itoh’s theorem \cite{itoh1982analysis}, as it does not require the assumption that intensity differences between neighboring pixels remain within half a period. Furthermore, as shown in \Fref{fig: LAR} (c) and (d), this equivalence extends to the Laplacian operator $\Delta$ and remains valid after applying a Poisson solver $\mathcal{P}$:
\begin{equation}
    \begin{dcases}
    \mathcal{R}(\Delta \mathbf{I}_m) = \mathcal{R}(\Delta \mathbf{I})\\
    \mathcal{P}(\mathcal{R}(\Delta \mathbf{I}_m)) = \mathcal{P}(\mathcal{R}(\Delta \mathbf{I}))
    \end{dcases}
    .
\end{equation}
Intuitively, $\mathcal{R}(\nabla \mathbf{I}_m)$ preserves high-frequency content of the underlying HDR scene, while $\mathcal{P}(\mathcal{R}(\Delta \mathbf{I}_m))$ captures complementary low-frequency content.

\textbf{Pyramidal multi-frequency adapter (PMF-Adapter).} Motivated by the above observation, we design the PMF-Adapter to bridge modulo-encoded measurements and pre-trained diffusion models. It addresses two key challenges: effectively integrating complementary information distributed across different frequency domains, and aligning such information with the inherently multiscale nature of diffusion-based denoising. As shown in the upper part of \Fref{fig: Algorithm} (a), the PMF-Adapter comprises two components: a multi-frequency aggregation module $\mathcal{F}_{\mathrm{fa}}$ and a pyramidal fusion module $\mathcal{F}_{\mathrm{pm}}$. Specifically, $\mathcal{F}_{\mathrm{fa}}$ extracts and fuses features from the modulo image $\mathbf{I}_m$ together with its LAR-processed gradient and Poisson-reconstructed Laplacian representations, \ie, $\mathcal{R}(\nabla \mathbf{I}_m)$ and $\mathcal{P}(\mathcal{R}(\Delta \mathbf{I}_m))$. These representations capture complementary structural information across frequency bands and are adaptively integrated using spatial-channel dual attention mechanisms implemented via CBAM blocks \cite{woo2018cbam}. Subsequently, $\mathcal{F}_{\mathrm{pm}}$ organizes the aggregated features into a hierarchical representation through scale-specific transformer layers \cite{dosovitskiy2020image}, producing a set of multiscale features $\mathbf{F}_{1,2,3,4}^a$. The overall workflow of the PMF-Adapter can be expressed as:
\begin{equation}
    \mathbf{F}_{1,2,3,4}^a = \mathcal{F}_{\mathrm{pm}} \big(\mathcal{F}_{\mathrm{fa}}(\mathbf{I}_m, \mathcal{R}(\nabla \mathbf{I}_m), \mathcal{P}(\mathcal{R}(\Delta \mathbf{I}_m))) \big).
\end{equation}
These features are then used to drive a ControlNet-like control injection module $\mathcal{C}$, enabling HDR prior extraction within a pre-trained diffusion model (Stable Diffusion v1.5 \cite{rombach2022high}).

\subsection{Stage2: Physics-Consistent Modulo Unwrapping} 
As shown in \Fref{fig: Algorithm} (b), this stage leverages the extracted HDR prior, represented by the latent code $\mathbf{z}_0$ obtained from the diffusion model, to perform modulo unwrapping. Instead of predicting the linear HDR image $\mathbf{I}$, we choose to reconstruct the $\mu$-law tone-mapped HDR image, which is defined as: 
\begin{equation}
    \label{eq: Mu}
    \mathbf{I}_\mu = \frac{\log(1 + \mu \mathbf{I})}{\log(1 + \mu)}.
\end{equation}
We adopt this logarithmic mapping because $\mathbf{I}_\mu$ better aligns with the distribution learned by the pre-trained diffusion model. To achieve this goal, a naive solution is to directly retrain a variational autoencoder (VAE) decoder using paired samples $(\mathbf{z}_0, \mathbf{I}_\mu)$. However, since $\mathbf{z}_0$ is generated by a pre-trained diffusion model under strong natural image priors, such a standalone decoder often suffers from distribution mismatch or fails to fully exploit the structural information encoded in the modulo measurements, resulting in unstable reconstruction, particularly in severely rolled-over regions.

\textbf{Latent-modulated attention decoder (LMA-Decoder).} To address the above limitation, we design the LMA-Decoder, which extends the original VAE decoder by explicitly introducing multiscale features from the PMF-Adapter (\ie, $\mathbf{F}_{1,2,3,4}^a$) to regulate the decoding process. As
shown in the left part of \Fref{fig: Algorithm} (b), instead of naively concatenating $\mathbf{F}_{1,2,3,4}^a$ with the decoder features $\mathbf{F}_{1,2,3,4}^d$ at each scale, we introduce an attention modulation block (AMB) to enable more precise, content-adaptive feature modulation. Specifically, as depicted in \Fref{fig: Algorithm} (c), at the $i$-th scale, the corresponding AMB jointly takes the decoder feature $\mathbf{F}_i^d$ and the adapter feature $\mathbf{F}_i^a$ as input. It predicts an offset $\mathbf{O}_i$ and a modulator $\mathbf{M}_i$ through a series of convolution layers and residual blocks \cite{he2016deep} equipped with instance normalization \cite{ulyanov2016instance}. The predicted $\mathbf{O}_i$ and $\mathbf{M}_i$ are used to parameterize a deformable convolution layer \cite{zhu2019deformable}, allowing the decoder features to be adaptively realigned, thereby mitigating spatial distortions. The resulting features are further transformed into an attention map $\mathbf{A}_i$, which modulates $\mathbf{F}_i^d$ in a residual manner to produce the updated decoder feature $\mathbf{F}_i^{d+}$. The updated feature $\mathbf{F}_i^{d+}$ directly replaces the original one $\mathbf{F}_i^d$ and is propagated to subsequent layers of the VAE decoder, enabling the modulation to be seamlessly integrated into the original decoding pipeline. Overall, the workflow of the LMA-Decoder can be summarized as:
\begin{equation}
    \mathbf{F}_{i-1}^{d} = \mathcal{D}_i \big( \text{AMB}(\mathbf{F}_i^{d}, \mathbf{F}_i^{a}) \big), \quad i = 4,3,2,1,
\end{equation}
where $\mathcal{D}_i$ denotes the $i$-th decoding operation of the VAE decoder. This design enables fine-grained measurement-driven details to effectively complement the global generative prior during reconstruction. 

\textbf{Cyclic-consistent physical refiner (CCP-Refiner).} Nevertheless, the VAE-style decoding only produces a coarse intermediate estimate $\hat{\mathbf{I}}_\mu$. Although perceptually plausible, this intermediate output may violate the deterministic physical constraints of modulo imaging when mapped back to the linear domain. Due to the periodic ambiguity introduced by the modulo operation, directly refining the intermediate estimate $\hat{\mathbf{I}}_\mu$ in the $\mu$-law tone-mapped domain often leads to physically inconsistent results in the linear irradiance domain, particularly near rollover boundaries. To address this issue, we design the CCP-Refiner to enforce physical consistency in a post-processing manner, explicitly regularizing the reconstruction across both the $\mu$-law tone-mapped and linear domains. As shown in the right part of \Fref{fig: Algorithm} (b), we first linearize $\hat{\mathbf{I}}_\mu$ to obtain a linear-domain estimate $\hat{\mathbf{I}}$. We then introduce a cyclic encoding that embeds $\hat{\mathbf{I}}$ into a sinusoidal space, enabling the refiner to explicitly reason about modulo periodicity. This design is highly consistent with the intrinsic periodic structure of modulo-based HDR imaging, where irradiance values differing by integer multiples of the modulo period are physically indistinguishable. Specifically, the cyclic encoding is defined as:
\begin{equation}
    \mathcal{E}(\hat{\mathbf{I}}) = \left\{ \sin(2\pi \boldsymbol{\phi}), \cos(2\pi \boldsymbol{\phi}) \right\}, \boldsymbol{\phi} = \frac{\text{mod}(\hat{\mathbf{I}}, 2^N)}{2^N}.
\end{equation}
Subsequently, we employ an autoencoder backbone $\mathcal{B}$ \cite{hinton2006reducing} to perform multiscale refinement on $\hat{\mathbf{I}}_\mu$, guided by the raw modulo measurement $\mathbf{I}_m$ and the cyclic embedding $\mathcal{E}(\hat{\mathbf{I}})$. The overall workflow of the CCP-Refiner is formulated as:
\begin{equation}
    \mathbf{I}_\mu = \mathcal{B}(\hat{\mathbf{I}}_\mu, \mathbf{I}_m, \mathcal{E}(\hat{\mathbf{I}})).
\end{equation}
Through this cyclic-guided refinement, the CCP-Refiner successfully corrects physically inconsistent components while preserving perceptual quality, yielding the final, physically faithful reconstruction $\mathbf{I}_\mu$. The corresponding linear HDR image $\mathbf{I}$ is then readily recovered by applying the inverse transformation of \Eref{eq: Mu}.

\subsection{Overall Implementation Details}
\textbf{Loss function for Stage1.} The loss function for Stage1 follows the standard denoising objective of diffusion models. Specifically, we minimize the mean squared error between the ground-truth noise (denoted as $\boldsymbol{\epsilon}$) and the noise predicted by the denoising network (denoted as $\boldsymbol{\epsilon}_\theta$):
\begin{equation}
    \mathcal{L}_\text{diff} = \left\| \boldsymbol{\epsilon} - \boldsymbol{\epsilon}_\theta \big( \mathbf{z}_d, d, \mathcal{C}(\mathbf{F}_{1,2,3,4}^a) \big) \right \|_2^2,
\end{equation}
where $\mathbf{z}_d$ denotes the noisy latent code at diffusion step $d$, and $\mathcal{C}(\mathbf{F}_{1,2,3,4}^a)$ represents the control features from the PMF-Adapter injected via the control injection module $\mathcal{C}$.

\textbf{Loss functions for Stage2.} The optimization objective for Stage2 consists of a decoder loss $\mathcal{L}_\text{dec}$ and a refiner loss $\mathcal{L}_\text{ref}$, which supervise the outputs of the LMA-Decoder and the CCP-Refiner, respectively. Using the superscript $\text{gt}$ to denote the ground truth, $\mathcal{L}_\text{dec}$ is defined as:
\begin{equation}
    \begin{split}
        \mathcal{L}_\text{dec}(\hat{\mathbf{I}}_\mu, \hat{\mathbf{I}}, \mathbf{I}_\mu^\text{gt}, \mathbf{I}^\text{gt}) &= \lambda_1 \| \hat{\mathbf{I}}_\mu - \mathbf{I}_\mu^\text{gt} \|_1 + \lambda_2 \| \hat{\mathbf{I}} - \mathbf{I}^\text{gt} \|_1 \\
        &+ \lambda_3 L_p(\hat{\mathbf{I}}_\mu, \mathbf{I}_\mu^\text{gt}),
    \end{split}
\end{equation}
where $\hat{\mathbf{I}}_\mu$ and $\hat{\mathbf{I}}$ are the intermediate reconstructions in the $\mu$-law tone-mapped and linear domains, respectively, $L_p$ denotes the perceptual loss, and the weighting coefficients are empirically set to $\lambda_1 = 10.0$, $\lambda_2 = 10.0$, and $\lambda_3 = 0.1$. The refiner loss $\mathcal{L}_\text{ref}$ builds upon $\mathcal{L}_\text{dec}$ by introducing self-supervised regularization terms in the linear domain to explicitly enforce physical consistency across multiple differential orders:
\begin{equation}
    \begin{split}
        \mathcal{L}_\text{ref}(\mathbf{I}_\mu, \mathbf{I}, \mathbf{I}_\mu^\text{gt}, \mathbf{I}^\text{gt}, \mathbf{I}_m) &= \mathcal{L}_\text{dec}(\mathbf{I}_\mu, \mathbf{I}, \mathbf{I}_\mu^\text{gt}, \mathbf{I}^\text{gt}) + L_\text{mod}(\mathbf{I}, \mathbf{I}_m) \\
        &+ L_\text{grad}(\mathbf{I}, \mathbf{I}_m) + L_\text{lap}(\mathbf{I},\mathbf{I}_m),
    \end{split}
\end{equation}
where $\mathbf{I}_\mu$ and $\mathbf{I}$ denote the final reconstructions in the $\mu$-law tone-mapped and linear domains, respectively. Here, the terms $L_\text{mod}$, $L_\text{grad}$, and $L_\text{lap}$ strictly enforce zeroth-, first-, and second-order physical consistency with the raw measurement $\mathbf{I}_m$, respectively. They are formulated as:
\begin{equation}
    \begin{dcases}
        L_\text{mod}(\mathbf{I}, \mathbf{I}_m) = \| \mathcal{R}(\mathbf{I} - \mathbf{I}_m) \|_1 \\
        L_\text{grad}(\mathbf{I}, \mathbf{I}_m) = \| \mathcal{R}(\nabla \mathbf{I}) - \mathcal{R}(\nabla \mathbf{I}_m) \|_1 \\
        L_\text{lap}(\mathbf{I}, \mathbf{I}_m) = \| \mathcal{R}(\Delta \mathbf{I}) - \mathcal{R}(\Delta \mathbf{I}_m) \|_1
    \end{dcases}.
\end{equation}

\textbf{Training protocol.} Our method is implemented in PyTorch and trained on four NVIDIA RTX 4090 GPUs. The training proceeds progressively to ensure stability. In Stage1, we train the PMF-Adapter and the control injection module $\mathcal{C}$ while keeping the pre-trained denoising network $\boldsymbol{\epsilon}_\theta$ strictly frozen. We adopt DDIM sampling with a linear noise schedule, setting the base learning rate to $1\times10^{-5}$, and train this stage for $2000$ epochs. After Stage1 converges, all its parameters are entirely frozen, and Stage2 training commences. In Stage2, we first train the LMA-Decoder independently for $2000$ epochs with a learning rate of $1\times10^{-4}$. Subsequently, the LMA-Decoder is fixed, and only the CCP-Refiner is optimized for another $2000$ epochs using the same learning rate.

\textbf{Optimization details.} All stages are optimized using the AdamW optimizer \cite{loshchilov2017decoupled} with standard momentum parameters ($\beta_1=0.9$, $\beta_2=0.999$) and a weight decay of $1\times10^{-4}$. We employ a cosine annealing learning rate schedule with a linear warmup \cite{loshchilov2016sgdr} throughout the entire training process. The batch size is uniformly set to $4$ for all experiments. Additionally, for the $\mu$-law tone mapping in Stage2, we fix the compression parameter to $\mu=5000$.

\begin{figure*}[t]
  \centering
  \includegraphics[width=\linewidth]{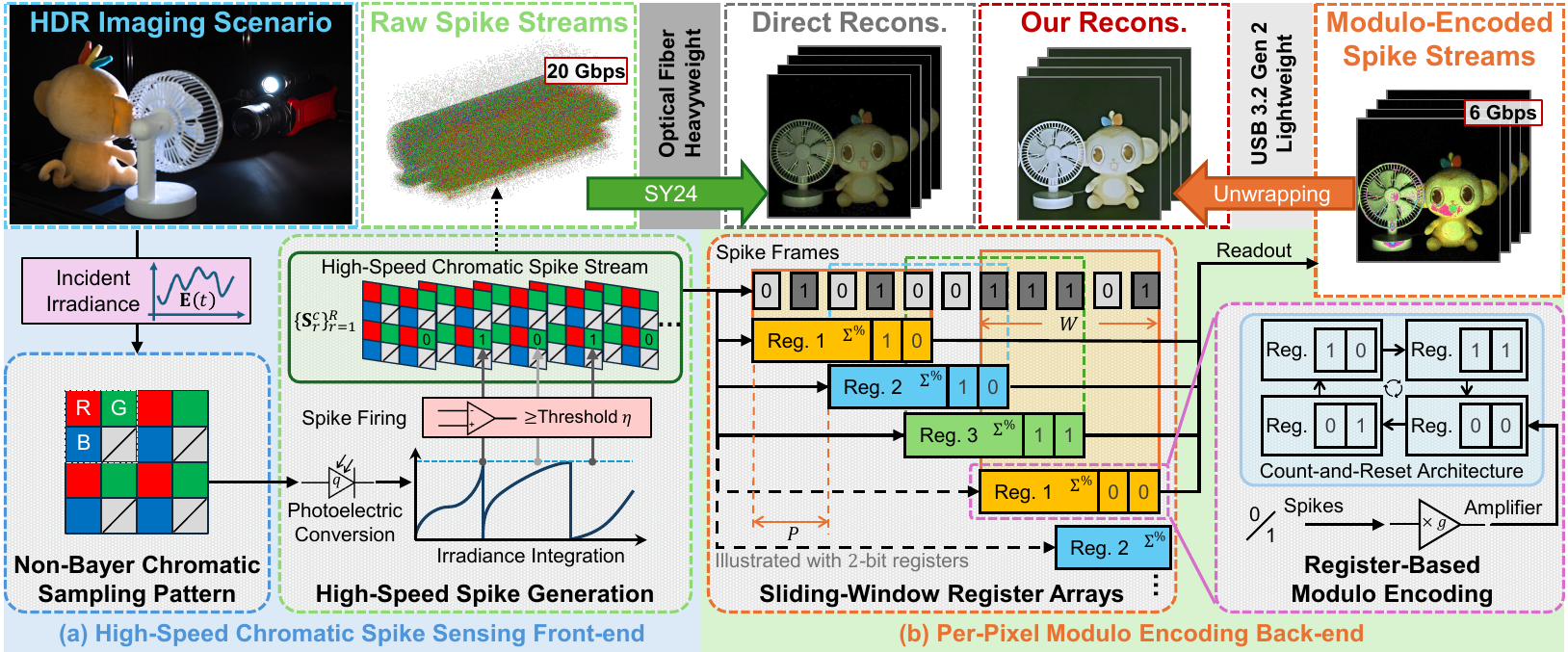}
  \caption{Illustration of the proposed bandwidth-efficient hardware implementation, designed as a proof-of-concept to validate the practical viability of our modulo imaging system. The architecture comprises two key components: (a) a high-speed chromatic spike sensing front-end, which utilizes a non-Bayer sampling pattern to physically instantiate the representation phase by generating a temporally dense chromatic spike stream; and (b) a per-pixel modulo encoding back-end, which executes the query phase through sliding-window register arrays, per-pixel amplification, and a count-and-reset mechanism. Compared to conventional spike-based HDR pipelines (\eg, SY24 \cite{yang2024real}) that must transmit raw, high-rate spike streams, our modulo encoding successfully reduces the output bandwidth from approximately $20$ Gbps to $6$ Gbps, enabling practical transmission via a standard USB 3.2 Gen 2 interface rather than requiring bulky optical fiber connections.}
  \label{fig: Sensor}
\end{figure*}

\section{Bandwidth-Efficient Hardware Implementation}
Implementing our exposure-decoupled formulation in physical hardware requires capturing the incident irradiance $\mathbf{E}(t)$ at an extremely high temporal resolution for each pixel. In practice, streaming such temporally dense measurements inevitably incurs a prohibitive bandwidth bottleneck. Spike cameras, which asynchronously emit binary spikes with a temporal density proportional to the underlying irradiance, offer a natural and highly efficient integral representation of $\mathbf{E}(t)$. This characteristic closely aligns with the mathematical requirements of our formulation's representation phase. Rather than fabricating a custom silicon sensor from scratch, our primary goal in this section is to demonstrate the practical viability of our proposed imaging system through a proof-of-concept hardware implementation. To this end, we physically reconfigure an off-the-shelf spike camera to serve as the foundation of our design. As illustrated in \Fref{fig: Sensor}, our prototype architecture comprises two primary components: (a) a high-speed chromatic spike sensing front-end and (b) a per-pixel modulo encoding back-end. By physically instantiating the representation and query phases of our formulation, this prototype successfully bridges the gap between theoretical modeling and physical realization, enabling bandwidth-efficient, high-speed, and full-color modulo imaging without the need for bespoke silicon fabrication.

\subsection{High-Speed Chromatic Spike Sensing Front-end}
As shown in \Fref{fig: Sensor} (a), the front-end is designed to physically instantiate the representation phase of our exposure-decoupled formulation. However, directly obtaining all short-time irradiance integrals $\{\mathbf{U}_k\}_{k=1}^{K}$ over the total capture duration $T$ requires a prohibitively large transmission bandwidth and massive storage, rendering it entirely impractical for hardware implementation. To circumvent this hardware barrier, we observe that spike cameras \cite{zhu2019retina} provide a natural and bandwidth-efficient alternative. A spike camera employs a retina-inspired integrate-and-fire mechanism, wherein each pixel continuously integrates the incident irradiance $\mathbf{E}(t)$. Whenever the accumulated irradiance since the last reset time $t_\text{last}$ reaches a predefined threshold $\eta$, a spike is emitted. Mathematically, the firing condition at time $t$ is triggered when:
\begin{equation}
    \int_{t_\text{last}}^{t} q \mathbf{E}(\tau) \mathrm{d}\tau \ge \eta,
\end{equation}
where $q$ denotes the photoelectric conversion gain. Upon firing, the integrator is immediately reset. As a result, each discrete spike corresponds to the accumulation of a fixed quantum of irradiance (up to the constant gain $q$), naturally yielding a sparse, event-driven integral representation of the scene dynamics. 

Although spike firing is inherently asynchronous at the physical level, the sensor state is synchronously read out at a fixed rate $f$. This process partitions the total capture duration $T$ into $R = fT$ uniform readout intervals. At each readout instant, a binary value is recorded per pixel, indicating whether at least one spike has occurred within that specific interval. This produces a sequence of binary spike frames $\{\mathbf{S}_r\}_{r=1}^{R} \in \{0,1\}^{R \times h \times w}$, which together form the discrete spike stream output. In practice, the readout rate is sufficiently high (typically $f \geq 20$ kHz) such that multiple firings within a single micro-interval are exceedingly rare, ensuring high-fidelity temporal capture. Under this high-speed regime, summing the short-time irradiance integrals over a specific temporal window can be robustly approximated by counting the number of spikes within the corresponding readout intervals, scaled by $\eta/q$:
\begin{equation}
    \label{eq: SpikeCount}
    \sum_{k=b_i}^{b_i + L - 1} \mathbf{U}_k \approx \frac{\eta}{q} \sum_{r \in \mathcal{R}(i)} \mathbf{S}_r, 
\end{equation}
where the index set $\mathcal{R}(i)$ is exactly defined to align with the continuous exposure duration:
\begin{equation}
    \mathcal{R}(i) = \left\{ r \in \mathbb{Z} \;\Big|\; b_i - 1 < r\frac{K}{R} \le b_i + L - 1 \right\}.
\end{equation}
Here, $\mathcal{R}(i)$ accurately indexes the spike frames whose readout intervals overlap with the precise physical duration covered by $\{\mathbf{U}_k\}_{k=b_i}^{b_i+L-1}$. Consequently, our mathematical formulation and the physical spike domains are rigorously linked by a temporal consistency condition:
\begin{equation}
    \frac{L}{K} = \frac{|\mathcal{R}(i)|}{R},
\end{equation}
which guarantees that both sides of \Eref{eq: SpikeCount} integrate the irradiance over an identical temporal duration. This equivalence establishes binary spike frames as a hardware-level surrogate for the representation phase in our formulation, entirely eliminating the need to explicitly record and transmit $\{\mathbf{U}_k\}_{k=1}^{K}$.

Despite these architectural advantages, incorporating chromatic information into spike-based modulo imaging introduces a fundamental conflict. Standard spike-based vision systems typically rely on a Bayer color filter array \cite{yang2024real}. However, we find that the spatial mosaic structure of Bayer sampling is intrinsically incompatible with the modulo paradigm. Specifically, the spatial interleaving of Bayer channels introduces severe modulo-induced discontinuities between neighboring pixels. Applying conventional demosaicing algorithms across these sharp discontinuities results in irrecoverable reconstruction artifacts, as spatial interpolation inevitably mixes samples from completely different wrapping states. This fundamentally prevents the accurate recovery of the per-pixel RGB measurements required for modulo unwrapping. To address this critical issue, we equip our hardware prototype with a non-Bayer chromatic sampling pattern explicitly tailored for spike-based modulo imaging. Concretely, we modify the standard $2 \times 2$ Bayer layout by omitting one green filter, yielding an effective macro-pixel block that contains exactly one red, one green, and one blue filtered pixel. By treating these three measurements within each block as approximately co-located observations, we entirely bypass the flawed traditional demosaicing process and directly construct spatially aligned RGB channels. The resulting chromatic spike stream is formalized as:
\begin{equation}
    \left\{\mathbf{S}_r^{(c)}\right\}_{r=1}^{R} \in \{0,1\}^{R \times \frac{h}{2} \times \frac{w}{2} \times 3}, \quad c \in \{1,2,3\},
\end{equation}
where $c$ indexes the color channels. This spatially aligned chromatic spike stream serves as the exact input to the subsequent query phase.

\subsection{Per-Pixel Modulo Encoding Back-end}
As illustrated in \Fref{fig: Sensor} (b), the back-end is designed to physically execute the query phase of our exposure-decoupled formulation. Building upon the temporal equivalence established in \Eref{eq: SpikeCount}, the abstract query operation defined in \Eref{eq: OurFormulation} can be practically reduced to per-pixel spike counting (scaled by an amplification factor $g = \eta h/q$) followed by a digital modulo operation. To realize this, we implement a per-pixel, fully parallel count-and-reset architecture utilizing dedicated register arrays. For each pixel and color channel, the back-end maintains multiple registers that operate on sliding temporal windows over the incoming chromatic spike stream $\{\mathbf{S}_r^{(c)}\}_{r=1}^{R} \in \{0,1\}^{R \times \frac{h}{2} \times \frac{w}{2} \times 3}$. Each window spans $W$ consecutive spike frames and advances with a fixed temporal stride $P$, acting as the hardware-domain discrete counterparts to the theoretical query window length and stride. To align the spike-domain counting with our ideal formulation, the accumulated spike count is passed through a per-pixel amplifier with an amplification factor $g = \eta h/q$. This sliding-window and amplification design is essential to ensure strict temporal and radiometric correspondence between the abstract query operation and its physical spike-domain implementation.

To align exactly with the 1-based indexing of the incoming spike stream, the starting index for the $j$-th query window is set to $a_j = (j-1)P + 1$. Formally, for each color channel $c$, the $j$-th modulo image produced by the encoding back-end is formulated as:
\begin{equation}
    \mathbf{I}_m^{(j,c)} = \text{mod} \left( \left\lfloor g \sum_{r=a_j}^{a_j+W-1} \mathbf{S}_r^{(c)} \right\rfloor, 2^N \right).
\end{equation}
By performing spike counting, amplifying, and modulo encoding entirely at the pixel level, this back-end architecture achieves massive parallelism and high bandwidth efficiency. Importantly, it natively outputs standard modulo images that are directly compatible with existing modulo unwrapping algorithms, allowing seamless integration with our downstream diffusion-based HDR reconstruction pipeline.

\textbf{Bandwidth reduction analysis.} Beyond merely serving as a physical surrogate for our theoretical formulation, the proposed encoding back-end fundamentally resolves the data transmission bottleneck inherent to high-speed spike-based imaging. In conventional spike-based HDR pipelines like SY24 \cite{yang2024real}, the sensor outputs binary spike frames at a fixed readout rate $f$, and existing reconstruction methods typically mandate continuous access to this raw spike stream. In practice, this stream is commonly produced using a Bayer pattern, yielding a single-channel spike sequence at full spatial resolution (\eg, $1000 \times 1000 \times 1$), which must be transmitted across the primary data interface before demosaicing. At a typical readout rate of $f = 20$ kHz, transmitting such a full-resolution spike stream requires a prohibitive bandwidth of $1000 \times 1000 \times 1 \times 20,000 \approx 20$ Gbps (2.5 GB/s). In contrast, our system adopts a non-Bayer chromatic spike representation with a resolution of $500 \times 500 \times 3$, performs temporal aggregation and modulo encoding internally, and outputs an $N$-bit modulo image only once every $P$ spike frames. Assuming an 8-bit depth ($N = 8$) and a stride of $P = 20$, the resulting output rate drops drastically to $f/P = 1$ kHz (achieving 1000 FPS capture). This corresponds to an output bandwidth of only $500 \times 500 \times 3 \times 8 \times 1000 \approx 6$ Gbps (0.75 GB/s), which comfortably fits within the capacity of a standard USB 3.2 Gen 2 protocol. Consequently, while conventional spike-based methods rely on flooding the transmission interface with high-rate raw spikes—often necessitating expensive and bulky optical fiber hardware—our approach achieves a substantial 70\% reduction in transmission overhead. It successfully emits bandwidth-efficient, temporally aggregated modulo measurements without sacrificing the underlying high-speed irradiance information.

\begin{table*}[t]
    \centering
    \caption{Quantitative comparisons on synthetic data using the UnModNet dataset \cite{zhou2020unmodnet} between our algorithm and state-of-the-art HDR reconstruction approaches, including two modulo unwrapping methods (UnModNet \cite{zhou2020unmodnet} and PnP-UA \cite{bacca2024deep}) and two LDR-to-HDR methods (IntrinsicHDR \cite{dille2024intrinsic} and LEDiff \cite{wang2025lediff}).}
    \label{tab: SyntheticData}
    \begin{tabular}{lcccccc}
        \toprule
        Method & PSNR-L $\uparrow$ & SSIM-L $\uparrow$ & HDR-VDP-3 $\uparrow$ & PSNR-PU $\uparrow$ & SSIM-PU $\uparrow$ & Total Inference Time (s) $\downarrow$ \\
        \midrule
        Ours & \textbf{39.17} & \textbf{0.977} & \textbf{8.837} & \textbf{33.77} & \textbf{0.974} & \textbf{42.57} \\
        UnModNet \cite{zhou2020unmodnet} & 35.46 & 0.972 & 8.687 & 32.59 & 0.973 & 169.55 \\
        PnP-UA \cite{bacca2024deep} & 28.58 & 0.707 & 4.399 & 16.77 & 0.738 & 53.42 \\
        IntrinsicHDR \cite{dille2024intrinsic} & 30.56 & 0.836 & 6.609 & 32.73 & 0.954 & 58.00 \\
        LEDiff \cite{wang2025lediff} & 29.11 & 0.640 & 6.519 & 25.46 & 0.742 & 994.79 \\
        \bottomrule
    \end{tabular}
\end{table*}

\subsection{Overall Implementation Details}
\textbf{Platform configuration.} The sensory foundation of our prototype is an off-the-shelf spike camera (Spike M1K40-H2-Gen3\footnote{\url{https://www.spikesee.com/product.html}}), with the raw spike readout rate configured to $f = 20$ kHz. To validate our exposure-decoupled formulation without incurring the prohibitive cost and extended lead time of custom silicon sensor fabrication, the proposed per-pixel modulo encoding back-end is physically emulated at the firmware level. Specifically, we developed a specialized firmware routine executing on the host computer to perform the parallel spike counting and modulo encoding in real time. For comprehensive experimental evaluation and baseline comparisons, this firmware is also configured to retain and forward the raw binary spike stream to the storage drive.

\textbf{Parameter settings and hardware scalability.} In our prototype experiments, the register sliding window span $W$ and stride $P$ are set to $25$ and $20$, respectively. Coupled with the $20$ kHz readout rate, this configuration yields an effective modulo output rate of $f/P = 1$ kHz, seamlessly enabling the system to operate at $1000$ FPS. The per-pixel amplification factor is empirically set to $g = 15$. It is crucial to note that these specific parameter choices—and consequently, the achievable quantization levels and dynamic range of the current prototype—are fundamentally bounded by the inherent physical constraints of the commercial spike camera (\eg, dark current, intrinsic noise floor, and baseline photoelectric sensitivity), rather than by our theoretical framework. To maintain a viable signal-to-noise ratio (SNR) using this off-the-shelf hardware, the temporal window length and amplification must be deliberately constrained. However, our proposed modulo imaging architecture is inherently scalable. Deploying this exact back-end design onto a bespoke CMOS image sensor with an optimized full-well capacity and finer quantization capabilities would transparently unlock exponentially higher dynamic range and imaging fidelity, fully realizing the theoretical limits of our formulation.

\begin{figure*}[t]
  \centering
  \includegraphics[width=\linewidth]{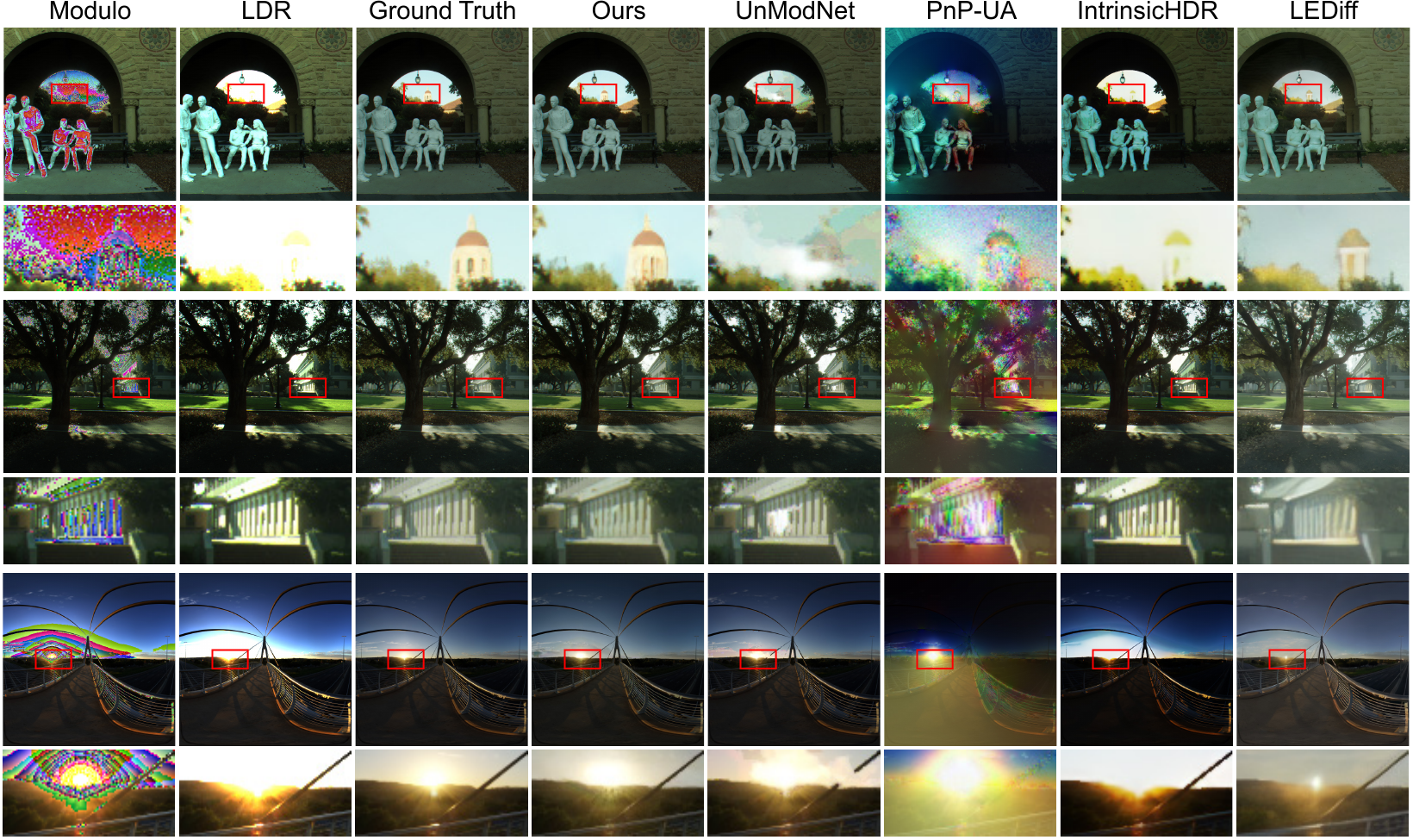}
  \caption{Qualitative comparisons on synthetic data using the UnModNet dataset \cite{zhou2020unmodnet} between our algorithm and state-of-the-art HDR reconstruction approaches, including two modulo unwrapping methods (UnModNet \cite{zhou2020unmodnet} and PnP-UA \cite{bacca2024deep}) and two LDR-to-HDR methods (IntrinsicHDR \cite{dille2024intrinsic} and LEDiff \cite{wang2025lediff}).}
  \label{fig: SyntheticData}
\end{figure*}

\section{Experiments}
\subsection{Evaluation on Synthetic Data}
\textbf{Datasets.} To rigorously evaluate the theoretical performance bound of our unwrapping algorithm and isolate its effectiveness from the physical imperfections of real-world hardware, we conduct comprehensive evaluations on the widely adopted UnModNet dataset \cite{zhou2020unmodnet}. For our experiments, we randomly partition its diverse scenes into 400 for training and 160 independent scenes for testing. The modulo sensor bit depth is configured to $N=8$, while the corresponding ground-truth HDR images are represented at a 12-bit depth. All synthetic images are processed at a fixed spatial resolution of $512 \times 512 \times 3$.

\textbf{Baselines.} We comprehensively benchmark our algorithm against two categories of state-of-the-art methods. The first category includes modulo unwrapping methods with publicly available code: UnModNet \cite{zhou2020unmodnet} and PnP-UA \cite{bacca2024deep}. The second category includes recent single-image LDR-to-HDR reconstruction methods, namely IntrinsicHDR \cite{dille2024intrinsic} and LEDiff \cite{wang2025lediff}. The inclusion of the latter highlights the fundamental advantages of modulo-based HDR imaging over conventional single-shot LDR-to-HDR paradigms.

\textbf{Evaluation metrics.} To comprehensively assess the reconstruction quality, we employ five widely adopted metrics spanning both physical accuracy and human perceptual fidelity. Specifically, we compute PSNR-L and SSIM-L in the linear radiance domain to measure absolute physical errors. To evaluate perceptual quality, which is crucial for HDR imaging, we report HDR-VDP-3 \cite{mantiuk2023hdr} alongside PSNR-PU and SSIM-PU, calculated after applying the perceptually uniform (PU21) encoding \cite{hanji2022comparison}. Furthermore, to benchmark computational efficiency, we measure the total inference time across the entire test set (consisting of 160 RGB images at a resolution of $512 \times 512 \times 3$) using a single NVIDIA RTX 4090 GPU.

\textbf{Quantitative and qualitative comparisons.} The quantitative results are summarized in \Tref{tab: SyntheticData}. Our proposed method consistently outperforms all competing baselines across all five physical and perceptual metrics. Notably, despite leveraging a powerful generative diffusion prior, our iteration-free architecture achieves the fastest inference speed among the evaluated methods, underscoring its exceptional computational efficiency. For qualitative evaluation, representative visual comparisons are provided in \Fref{fig: SyntheticData}\footnote{Additional results on synthetic data can be found in the PDF supplementary material.}. While existing single-shot LDR-to-HDR methods fail to hallucinate lost details in saturated areas, and prior modulo unwrapping algorithms struggle with severe structural distortions near dense rollover boundaries, our algorithm faithfully reconstructs high-fidelity textures and physically consistent gradients, closely resembling the ground-truth HDR scenes.

\begin{table*}[t]
    \centering
    \caption{Quantitative evaluation results of ablation study on synthetic data using the UnModNet dataset \cite{zhou2020unmodnet}.}
    \label{tab: AblationStudy}
    \begin{tabular}{lcccccc}
        \toprule
        Method & PSNR-L $\uparrow$ & SSIM-L $\uparrow$ & HDR-VDP-3 $\uparrow$ & PSNR-PU $\uparrow$ & SSIM-PU $\uparrow$\\
        \midrule
        Without Stage2 & 34.33 & 0.920 & 6.782 & 24.75 & 0.708 \\
        Without $\mathcal{R}(\nabla \mathbf{I}_m)$ and $\mathcal{P}(\mathcal{R}(\Delta \mathbf{I}_m))$          & 34.54 & 0.918 & 6.366 & 23.76 & 0.655 \\
        Direct control & 33.38 & 0.898 & 5,852 & 23.43 & 0.672 \\
        Without $\mathbf{F}_{1,2,3,4}^a$ & 36.28 & 0.950 & 6.953 & 25.99 & 0.748 \\
        Without AMBs & 37.94 & 0.970 & 8.035 & 29.45 & 0.926 \\
        Without CCP-Refiner & 38.64 & 0.975 & 8.793 & 33.48 & 0.972 \\
        Our complete algorithm & \textbf{39.17} & \textbf{0.977} & \textbf{8.837} & \textbf{33.77} & \textbf{0.974} \\
        \bottomrule
  \end{tabular}
\end{table*}

\begin{figure*}[t]
  \centering
  \includegraphics[width=\linewidth]{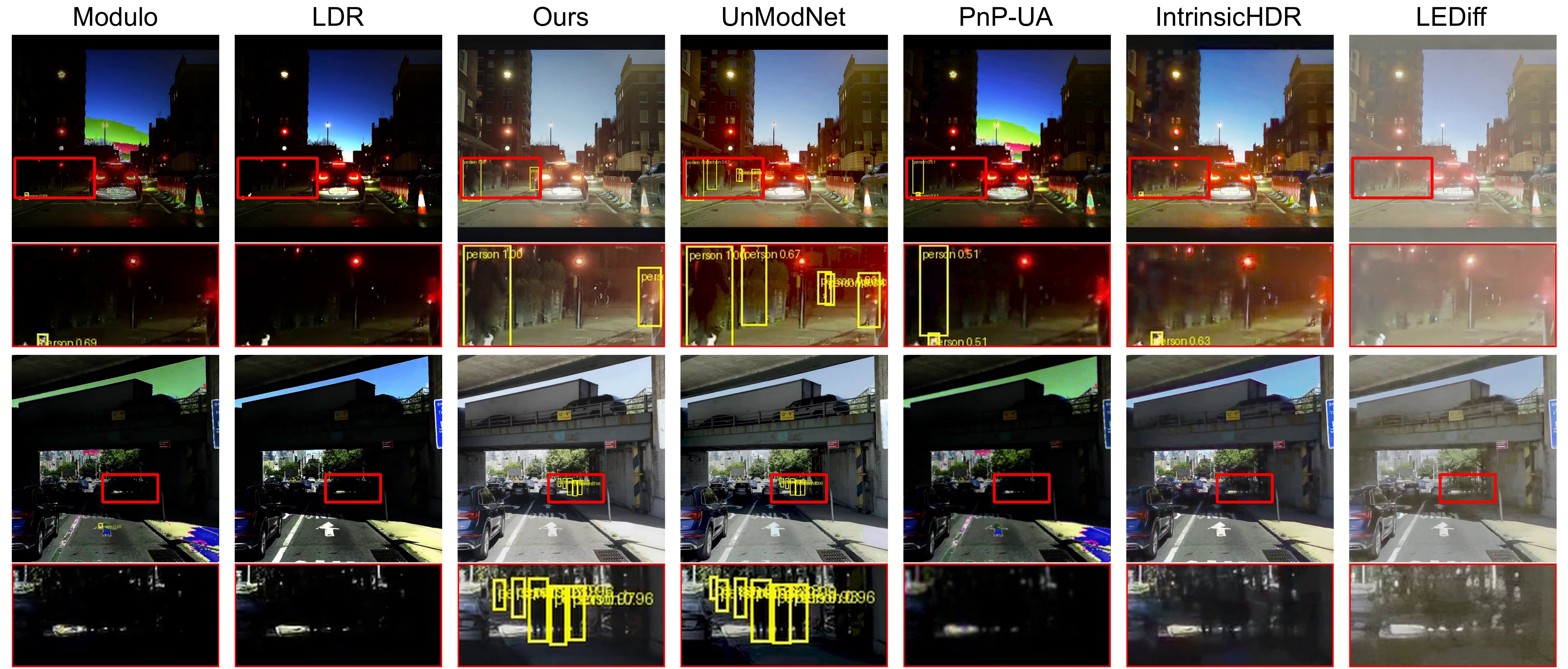}
  \caption{Comparisons on pedestrian detection using reconstructed HDR driving scenes between our algorithm and state-of-the-art HDR reconstruction approaches, including two modulo unwrapping methods (UnModNet \cite{zhou2020unmodnet} and PnP-UA \cite{bacca2024deep}) and two LDR-to-HDR methods (IntrinsicHDR \cite{dille2024intrinsic} and LEDiff \cite{wang2025lediff}). Detected pedestrian bounding boxes are overlaid in yellow for visualization.}
  \label{fig: Application}
\end{figure*}

\textbf{Ablation study.} To validate the efficacy of our core algorithmic designs, we conduct extensive ablation studies, with quantitative results reported in \Tref{tab: AblationStudy}. We systematically investigate four key aspects of our framework:
\begin{itemize}
    \item Two-stage architecture: We compare our full model with a single-stage variant (Without Stage2), where the diffusion latent code $\mathbf{z}_0$ is directly mapped to the image space using the original pre-trained VAE decoder. The severe performance drop highlights the absolute necessity of our measurement-guided Stage2 decoding.
    \item Multi-frequency LAR representations: We restrict the input to solely the raw modulo image $\mathbf{I}_m$ (Without $\mathcal{R}(\nabla \mathbf{I}_m)$ and $\mathcal{P}(\mathcal{R}(\Delta \mathbf{I}_m))$). The degraded metrics confirm the vital role of explicitly exploiting complementary high- and low-frequency structural priors derived from the LAR property.
    \item Feature modulation mechanisms: We assess three variants to validate the PMF-Adapter and LMA-Decoder. ``Direct control'' bypasses the hierarchical PMF-Adapter, injecting raw signals directly into the diffusion model. ``Without $\mathbf{F}_{1,2,3,4}^a$'' isolates the LMA-Decoder from the adapter's multiscale guidance. Finally, ``Without AMBs'' replaces our deformable attention modulation blocks with vanilla convolutions. The successive improvements brought by each component validate our precise, content-adaptive feature routing strategy.
    \item Physical consistency refinement: Removing the CCP-Refiner (Without CCP-Refiner) and treating the intermediate $\mu$-law estimate as the final output leads to sub-optimal linear-domain metrics, verifying the refiner's crucial role in explicitly reasoning about modulo periodicity and correcting deterministic physical violations.
\end{itemize}

\textbf{Downstream application.} To demonstrate the practical superiority of modulo-based HDR imaging and our specific unwrapping algorithm, we conduct a downstream pedestrian detection experiment on challenging driving scenes. We utilize a pre-trained \texttt{fasterrcnn\_resnet50\_fpn\_v2} model \cite{li2021benchmarking} implemented in PyTorch. Using HDR driving images sourced from public datasets, we simulate the corresponding sensor inputs and reconstruct the scenes using UnModNet, PnP-UA, IntrinsicHDR, LEDiff, and our method. As visualized in \Fref{fig: Application}, extreme contrast and severe overexposure severely degrade the performance of baseline methods, leading to missed detections or severe false positives caused by contrast loss. In contrast, our method successfully recovers structural details in previously washed-out regions, yielding highly robust and high-confidence detection results, thereby validating its immense potential for safety-critical downstream vision tasks.

\begin{figure*}[t]
  \centering
  \includegraphics[width=\linewidth]{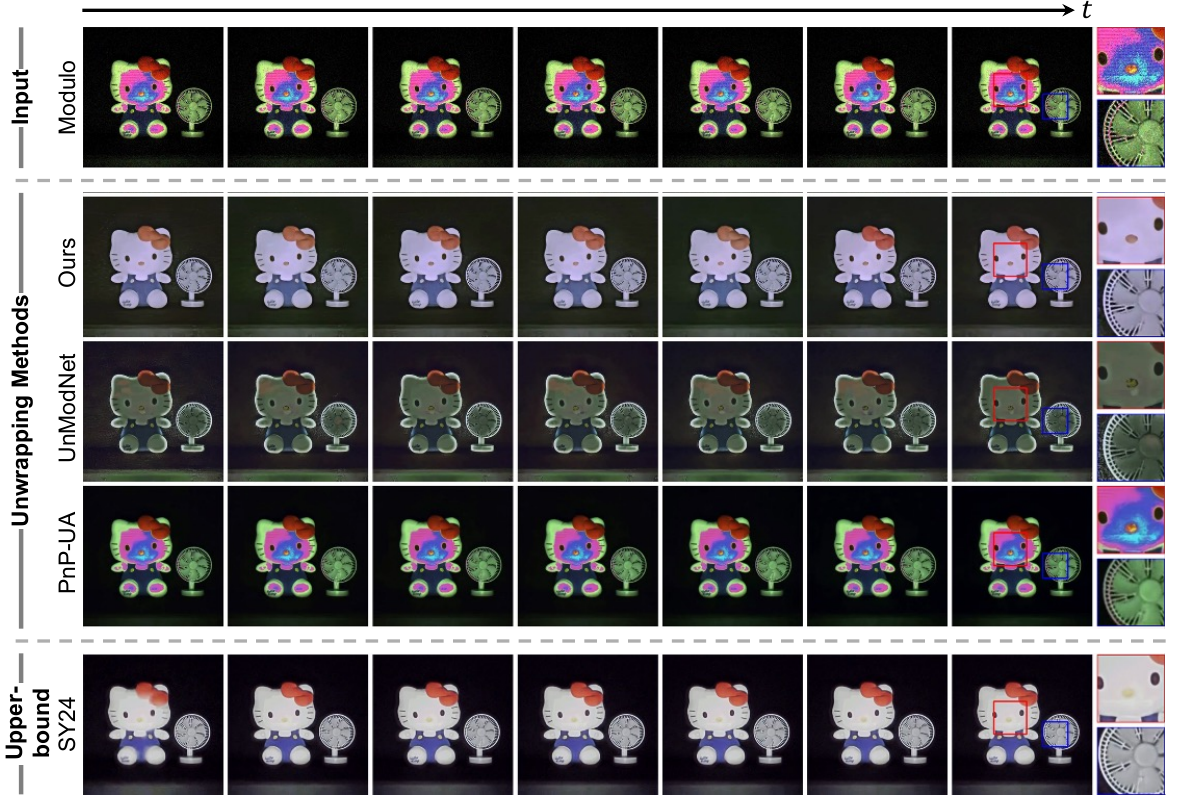}
  \caption{Qualitative comparisons on real-world dynamic scenes captured by our proof-of-concept hardware prototype. We compare our algorithm with two modulo-based HDR unwrapping approaches (UnModNet \cite{zhou2020unmodnet} and PnP-UA \cite{bacca2024deep}). Additionally, we include SY24 \cite{yang2024real}, a state-of-the-art conventional spike-based HDR video reconstruction method operating on complete raw spike streams, to serve as an information upper-bound baseline. Each image corresponds to a representative frame extracted from the reconstructed videos.}
  \label{fig: RealData}
\end{figure*}

\subsection{Evaluation on Real Hardware Data}
To validate the practical viability of our exposure-decoupled modulo imaging system, we deploy our proof-of-concept hardware prototype to capture real-world dynamic scenes. As a laboratory-stage prototype constructed from off-the-shelf components, its absolute dynamic range and light sensitivity are strictly bounded by the underlying sensor's physical constraints. Consequently, testing in unconstrained outdoor environments, where extreme illumination would easily exceed the sensor's physical operating limits, is not feasible. Instead, we conduct our evaluations in an indoor laboratory setting with moderate illumination contrast. This environment ensures the incident irradiance remains within the prototype's reliable response region, allowing us to successfully demonstrate the end-to-end functionality of the modulo imaging system without being bottlenecked by rudimentary hardware specifications. All physical experiments yield 1000 FPS full-color video streams at a resolution of $500 \times 500$, demonstrating high-speed acquisition under strict bandwidth constraints.

\textbf{Baselines and upper bounds.} We benchmark our real-world reconstructions against two modulo unwrapping algorithms (UnModNet \cite{zhou2020unmodnet} and PnP-UA \cite{bacca2024deep}), which serve as our direct and fair competitors. Furthermore, we include SY24 \cite{yang2024real}, a state-of-the-art conventional spike-based HDR video reconstruction method. It is crucial to note that SY24 operates under a fundamentally different paradigm: it necessitates the transmission of the complete, high-rate raw spike stream (consuming roughly $3\times$ our bandwidth) and employs heavily engineered, spike-specific denoising modules. Therefore, SY24 is included not as a direct peer, but as an \textit{information upper-bound baseline} to assess the imaging quality retained after our massive bandwidth reduction. To ensure a fair visual comparison under real-world sensor noise, a lightweight blind video denoiser \cite{lindner2023lightweight} is applied to the final outputs of all modulo-based methods, while SY24 relies on its built-in modules.

\textbf{Qualitative results and analysis.} Visual comparisons are presented in \Fref{fig: RealData}\footnote{Additional visual results on real-captured data, as well as the full dynamic video sequences, are provided in the supplementary materials.}, showcasing representative frames extracted from the reconstructed 1000 FPS videos. Our method significantly outperforms the competing modulo unwrapping algorithms, effectively suppressing real-world noise-induced artifacts and structural distortions that plague UnModNet and PnP-UA. Furthermore, despite utilizing only approximately $30\%$ of the raw sensory data, our framework achieves structural and textural fidelity highly comparable to the upper-bound baseline (SY24). We do observe two specific visual artifacts in our real-world results: minor global color distribution discrepancies compared to SY24, and slight inter-frame color fluctuations across the video sequence. The former arises because SY24 employs a perceptually tuned image signal processing (ISP) pipeline to produce visually ``pleasing'' colors, whereas our pipeline directly reconstructs the uncalibrated, raw sensory signals, prioritizing the mathematical preservation of the physical irradiance ratios. The latter occurs because we deliberately apply our algorithm to the video sequence frame-by-frame without temporal constraints. We maintain this independent spatial processing to ensure a rigorous and equitable comparison with existing modulo HDR baselines, which are all fundamentally designed for single-image reconstruction. Despite these specific artifacts, this hardware-in-the-loop experiment confirms that our imaging system successfully translates theoretical bandwidth efficiency into practical HDR acquisition without compromising essential structural fidelity.

\section{Conclusion}
We present a complete modulo-based HDR imaging system that enables high-speed, full-color HDR video acquisition. The proposed system comprises an exposure-decoupled formulation of modulo imaging, an iteration-free modulo unwrapping algorithm, and a bandwidth-efficient hardware implementation acting as a proof-of-concept prototype. Extensive experiments on synthetic benchmarks, downstream application scenarios, and real-captured data collectively demonstrate the practical viability of our system. It can operate robustly for $1000$ FPS HDR imaging with a $0.27$s unwrapping time per frame, while effectively reducing the transmission bandwidth to approximately $6$ Gbps.

\textbf{Limitations and future work.} The proposed unwrapping algorithm focuses on the modulo imaging formulation and does not explicitly address the demosaicing problem. As a result, our current proof-of-concept hardware prototype adopts a non-Bayer chromatic sampling pattern, which favors direct full-color acquisition at the cost of reduced spatial resolution. An important direction for future work is to develop a mosaicing-aware modulo unwrapping algorithm that is compatible with standard Bayer-pattern data, enabling higher-resolution reconstruction.

{\small
\bibliographystyle{ieee_fullname}
\bibliography{egbib}

@STRING{CVPR = {Proc. of Computer Vision and Pattern Recognition}}

@STRING{ECCV = {Proc. of European Conference on Computer Vision}}

@STRING{ICCV = {Proc. of International Conference on Computer Vision}}

@STRING{NIPS = {Proc. of Advances in Neural Information Processing Systems}}

@STRING{ICCP = {Proc. of International Conference on Computational Photography}}

@STRING{ICME = {Proc. of International Conference on Multimedia and Expo}}

@STRING{ICIP = {Proc. of International Conference on Image Processing}}

@STRING{WACV = {Proc. of Winter Conference on Applications of Computer Vision}}

@STRING{ICASSP = {Proc. of International Conference on Acoustics, Speech and Signal Processing}}

@STRING{TPAMI = {IEEE Transactions on Pattern Analysis and Machine Intelligence}}

@STRING{TIP = {IEEE Transactions on Image Processing}}

@STRING{IJCV = {International Journal of Computer Vision}}

@STRING{AO = {Applied Optics}}

@STRING{SIGGRAPH = {Proc. of ACM SIGGRAPH}}

@STRING{SIGGRAPHTOG = {ACM Transactions on Graphics (Proc. of ACM SIGGRAPH)}}

@STRING{SIGGRAPHTOGa = {ACM Transactions on Graphics (Proc. of ACM SIGGRAPH Asia)}}

@inproceedings{debevec1997recovering,
	title={Recovering high dynamic range radiance maps from photographs},
	author={Debevec, Paul E and Malik, Jitendra},
	booktitle=SIGGRAPH,
    pages={369--378},
	year={1997}
}

@inproceedings{khan2006ghost,
  title={Ghost removal in high dynamic range images},
  author={Khan, Erum Arif and Akyuz, Ahmet Oguz and Reinhard, Erik},
  booktitle=ICIP,
  pages={2005--2008},
  year={2006}
}

@article{sen2012robust,
    title={Robust patch-based {HDR} reconstruction of dynamic scenes},
	author={Sen, Pradeep and Kalantari, Nima Khademi and Yaesoubi, Maziar and Darabi, Soheil and Goldman, Dan B. and Shechtman, Eli},
	journal=SIGGRAPHTOG,
	volume={31},
	number={6},
	pages={203:1--203:11},
	year={2012}
}

@article{oh2014robust,
	title={Robust high dynamic range imaging by rank minimization},
	author={Oh, Tae-Hyun and Lee, Joon-Young and Tai, Yu-Wing and Kweon, In So},
	journal=TPAMI,
	volume={37},
	number={6},
	pages={1219--1232},
	year={2014}
}

@article{kalantari2017deep,
	title={Deep high dynamic range imaging of dynamic scenes},
	author={Kalantari, Nima Khademi and Ramamoorthi, Ravi},
	journal=SIGGRAPHTOG,
	volume={36},
	number={4},
	pages={144--1},
	year={2017}
}

@inproceedings{prabhakar2020towards,
    title={Towards practical and efficient high-resolution {HDR} deghosting with {CNN}},
    author={Prabhakar, K Ram and Agrawal, Susmit and Singh, Durgesh Kumar and Ashwath, Balraj and Babu, R Venkatesh},
    booktitle=ECCV,
    pages={497--513},
    year={2020}
}

@article{niu2021hdr,
    title={{HDR-GAN}: {HDR} image reconstruction from multi-exposed {LDR} images with large motions},
    author={Niu, Yuzhen and Wu, Jianbin and Liu, Wenxi and Guo, Wenzhong and Lau, Rynson WH},
    journal=TIP,
    volume={30},
    pages={3885--3896},
    year={2021}
}

@article{yan2022dual,
  title={Dual-attention-guided network for ghost-free high dynamic range imaging},
  author={Yan, Qingsen and Gong, Dong and Shi, Javen Qinfeng and Van Den Hengel, Anton and Shen, Chunhua and Reid, Ian and Zhang, Yanning},
  journal=IJCV,
  volume={130},
  number={1},
  pages={76--94},
  year={2022}
}

@inproceedings{hu2024generating,
  title={Generating content for {HDR} deghosting from frequency view},
  author={Hu, Tao and Yan, Qingsen and Qi, Yuankai and Zhang, Yanning},
  booktitle=CVPR,
  pages={25732--25741},
  year={2024}
}

@inproceedings{chen2025ultrafusion,
  title={{UltraFusion}: Ultra high dynamic imaging using exposure fusion},
  author={Chen, Zixuan and Wang, Yujin and Cai, Xin and You, Zhiyuan and Lu, Zheming and Zhang, Fan and Guo, Shi and Xue, Tianfan},
  booktitle=CVPR,
  pages={16111--16121},
  year={2025}
}

@inproceedings{li2025afunet,
  title={{AFUNet}: Cross-iterative alignment-fusion synergy for {HDR} reconstruction via deep unfolding paradigm},
  author={Li, Xinyue and Ni, Zhangkai and Yang, Wenhan},
  booktitle=ICCV,
  pages={10666-10675},
  year={2025}
}

@article{hasinoff2016burst,
	title={Burst photography for high dynamic range and low-light imaging on mobile cameras},
	author={Hasinoff, Samuel W and Sharlet, Dillon and Geiss, Ryan and Adams, Andrew and Barron, Jonathan T and Kainz, Florian and Chen, Jiawen and Levoy, Marc},
	journal=SIGGRAPHTOG,
	volume={35},
	number={6},
	pages={1--12},
	year={2016}
}

@inproceedings{liu2023joint,
  title={Joint {HDR} denoising and fusion: A real-world mobile {HDR} image dataset},
  author={Liu, Shuaizheng and Zhang, Xindong and Sun, Lingchen and Liang, Zhetong and Zeng, Hui and Zhang, Lei},
  booktitle=CVPR,
  pages={13966--13975},
  year={2023}
}

@inproceedings{kalantari2019deep,
  title={Deep {HDR} video from sequences with alternating exposures},
  author={Kalantari, Nima Khademi and Ramamoorthi, Ravi},
  booktitle={Computer Graphics Forum},
  volume={38},
  number={2},
  pages={193--205},
  year={2019}
}

@inproceedings{xu2024hdrflow,
  title={{HDRFlow}: Real-time {HDR} video reconstruction with large motions},
  author={Xu, Gangwei and Wang, Yujin and Gu, Jinwei and Xue, Tianfan and Yang, Xin},
  booktitle=CVPR,
  pages={24851--24860},
  year={2024}
}

@inproceedings{banterle2006inverse,
	title={Inverse tone mapping},
	author={Banterle, Francesco and Ledda, Patrick and Debattista, Kurt and Chalmers, Alan},
	booktitle={Proc. of International Conference on Computer Graphics and Interactive Techniques in Australasia and Southeast Asia},
	year={2006}
}

@article{masia2009evaluation,
    title={Evaluation of reverse tone mapping through varying exposure conditions},
	author={Masia, Belen and Agustin, Sandra and Fleming, Roland W. and Sorkine, Olga and Gutierrez, Diego},
	journal=SIGGRAPHTOGa,
	volume={28},
	number={5},
	pages={160:1--160:8},
	year={2009}
}

@article{rempel2007ldr2hdr,
    title={{LDR2HDR}: On-the-fly reverse tone mapping of legacy video and photographs},
	author={Rempel, Allan G. and Trentacoste, Matthew and Seetzen, Helge and Young, H. David and Heidrich, Wolfgang and Whitehead, Lorne and Ward, Greg},
	journal=SIGGRAPHTOG,
	volume={26},
	number={3},
	year={2007}
}

@article{endo2017deep,
    title={Deep reverse tone mapping},
	author={Endo, Yuki and Kanamori, Yoshihiro and Mitani, Jun},
	journal=SIGGRAPHTOGa,
	volume={36},
	number={6},
	pages={177:1--177:10},
	year={2017}
}

@article{eilertsen2017hdr,
    title={{HDR} image reconstruction from a single exposure using deep {CNNs}},
	author={Eilertsen, Gabriel and Kronander, Joel and Denes, Gyorgy and Mantiuk, Rafa\l K. and Unger, Jonas},
	journal=SIGGRAPHTOGa,
	volume={36},
	number={6},
	pages={178:1--178:15},
	year={2017}
}

@article{marnerides2018expandnet,
	title={{ExpandNet}: A deep convolutional neural network for high dynamic range expansion from low dynamic range content},
	author={Marnerides, Demetris and Bashford-Rogers, Thomas and Hatchett, Jonathan and Debattista, Kurt},
	journal={Computer Graphics Forum},
	volume={37},
	pages={37--49},
	year={2018}
}

@article{santos2020single,
    title={Single image {HDR} reconstruction using a {CNN} with masked features and perceptual loss},
    author={Santos, Marcel Santana and Ren, Tsang Ing and Kalantari, Nima Khademi},
    journal=SIGGRAPHTOG,
    volume={39},
    number={4},
    pages={80--1},
    year={2020}
}

@inproceedings{liu2020single,
	title={Single-image {HDR} reconstruction by learning to reverse the camera pipeline},
	author={Liu, Yu-Lun and Lai, Wei-Sheng and Chen, Yu-Sheng and Kao, Yi-Lung and Yang, Ming-Hsuan and Chuang, Yung-Yu and Huang, Jia-Bin},
	booktitle=CVPR,
	pages={1651--1660},
	year={2020}
}

@inproceedings{dille2024intrinsic,
  title={Intrinsic single-image {HDR} reconstruction},
  author={Dille, Sebastian and Careaga, Chris and Aksoy, Ya{\u{g}}{\i}z},
  booktitle=ECCV,
  pages={161--177},
  year={2024}
}

@inproceedings{wang2025lediff,
  title={{LEDiff}: Latent exposure diffusion for {HDR} generation},
  author={Wang, Chao and Xia, Zhihao and Leimkuhler, Thomas and Myszkowski, Karol and Zhang, Xuaner},
  booktitle=CVPR,
  pages={453--464},
  year={2025}
}

@inproceedings{nayar2000high,
	title={High dynamic range imaging: Spatially varying pixel exposures},
	author={Nayar, Shree K and Mitsunaga, Tomoo},
	booktitle=CVPR,
	pages={472--479},
	year={2000}
}

@inproceedings{schoberl2012high,
  title={High dynamic range video by spatially non-regular optical filtering},
  author={Sch{\"o}berl, Michael and Belz, Alexander and Seiler, J{\"u}rgen and Foessel, Siegfried and Kaup, Andr{\'e}},
  booktitle=ICIP,
  pages={2757--2760},
  year={2012}
}

@inproceedings{aguerrebere2014single,
  title={Single shot high dynamic range imaging using piecewise linear estimators},
  author={Aguerrebere, Cecilia and Almansa, Andr{\'e}s and Gousseau, Yann and Delon, Julie and Muse, Pablo},
  booktitle=ICCP,
  pages={1--10},
  year={2014}
}

@inproceedings{serrano2016convolutional,
  title={Convolutional sparse coding for high dynamic range imaging},
  author={Serrano, Ana and Heide, Felix and Gutierrez, Diego and Wetzstein, Gordon and Masia, Belen},
  booktitle={Computer Graphics Forum},
  pages={153--163},
  year={2016}
}

@inproceedings{alghamdi2019reconfigurable,
  title={Reconfigurable snapshot {HDR} imaging using coded masks and inception network},
  author={Alghamdi, Masheal M and Fu, Qiang and Thabet, Ali Kassem and Heidrich, Wolfgang},
  booktitle = {Vision, Modeling and Visualization},
  year={2019}
}

@inproceedings{metzler2020deep,
    title={Deep optics for single-shot high-dynamic-range imaging},
    author={Metzler, Christopher A and Ikoma, Hayato and Peng, Yifan and Wetzstein, Gordon},
    booktitle=CVPR,
    pages={1375--1385},
    year={2020}
}

@article{zhou2023polarizationHDR,
  title={Polarization guided {HDR} reconstruction via pixel-wise depolarization},
  author={Zhou, Chu and Han, Yufei and Teng, Minggui and Han, Jin and Li, Si and Xu, Chao and Shi, Boxin},
  journal=TIP,
  volume={32},
  pages={1774--1787},
  year={2023}
}

@article{brandli2014240,
    title={A 240 $\times$ 180 130 {dB} 3 $\mu$s latency global shutter spatiotemporal vision sensor},
    author={Brandli, Christian and Berner, Raphael and Yang, Minhao and Liu, Shih-Chii and Delbruck, Tobi},
    journal={IEEE Journal of Solid-State Circuits},
    volume={49},
    number={10},
    pages={2333--2341},
    year={2014}
}

@inproceedings{zhu2019retina,
	title={A retina-inspired sampling method for visual texture reconstruction},
	author={Zhu, Lin and Dong, Siwei and Huang, Tiejun and Tian, Yonghong},
	booktitle=ICME,
	pages={1432--1437},
	year={2019}
}

@article{rebecq2019high,
  title={High speed and high dynamic range video with an event camera},
  author={Rebecq, Henri and Ranftl, Ren{\'e} and Koltun, Vladlen and Scaramuzza, Davide},
  journal=TPAMI,
  volume={43},
  number={6},
  pages={1964--1980},
  year={2019}
}

@inproceedings{wang2021asynchronous,
    title={An asynchronous {Kalman} filter for hybrid event cameras},
    author={Wang, Ziwei and Ng, Yonhon and Scheerlinck, Cedric and Mahony, Robert},
    booktitle=ICCV,
    pages={448-457},
    year={2021}
}

@inproceedings{yang2024real,
  title={Real-data-driven 2000 {FPS} color video from mosaicked chromatic spikes},
  author={Yang, Siqi and Huang, Zhaojun and Chang, Yakun and Fan, Bin and Yu, Zhaofei and Shi, Boxin},
  booktitle=ECCV,
  pages={305--321},
  year={2024}
}

@inproceedings{han2020neuromorphic,
    title={Neuromorphic camera guided high dynamic range imaging},
    author={Han, Jin and Zhou, Chu and Duan, Peiqi and Tang, Yehui and Xu, Chang and Xu, Chao and Huang, Tiejun and Shi, Boxin},
    booktitle=CVPR,
    pages={1730--1739},
    year={2020}
}

@article{han2023hybrid,
  title={Hybrid high dynamic range imaging fusing neuromorphic and conventional images},
  author={Han, Jin and Yang, Yixin and Duan, Peiqi and Zhou, Chu and Ma, Lei and Xu, Chao and Huang, Tiejun and Sato, Imari and Shi, Boxin},
  journal=PAMI,
  year={2023}
}

@inproceedings{chang20231000,
  title={1000 {PFS} {HDR} video with a spike-{RGB} hybrid camera},
  author={Chang, Yakun and Zhou, Chu and Hong, Yuchen and Hu, Liwen and Xu, Chao and Huang, Tiejun and Shi, Boxin},
  booktitle=CVPR,
  pages={22180--22190},
  year={2023}
}

@inproceedings{zhao2015unbounded,
	title={Unbounded high dynamic range photography using a modulo camera},
	author={Zhao, Hang and Shi, Boxin and Fernandez-Cull, Christy and Yeung, Sai-Kit and Raskar, Ramesh},
	booktitle=ICCP,
	pages={1--10},
	year={2015}
}

@inproceedings{jagatap2020high,
  title={High dynamic range imaging using deep image priors},
  author={Jagatap, Gauri and Hegde, Chinmay},
  booktitle=ICASSP,
  pages={9289--9293},
  year={2020}
}

@inproceedings{zhou2020unmodnet,
    title={{UnModNet}: Learning to unwrap a modulo image for high dynamic range imaging},
    author={Zhou, Chu and Zhao, Hang and Han, Jin and Xu, Chang and Xu, Chao and Huang, Tiejun and Shi, Boxin},
    booktitle=NIPS,
    year={2020}
}

@inproceedings{bacca2024deep,
  title={Deep plug-and-play algorithm for unsaturated imaging},
  author={Bacca, Jorge and Monroy, Brayan and Arguello, Henry},
  booktitle=ICASSP,
  pages={2460--2464},
  year={2024}
}

@inproceedings{chen2025robust,
  title={Robust unfolding network for {HDR} imaging with modulo cameras},
  author={Chen, Zhile and Ji, Hui},
  booktitle=ICCV,
  pages={25218--25228},
  year={2025}
}

@inproceedings{ho2020denoising,
  title={Denoising diffusion probabilistic models},
  author={Ho, Jonathan and Jain, Ajay and Abbeel, Pieter},
  booktitle=NIPS,
  pages={6840--6851},
  year={2020}
}

@article{song2020denoising,
  title={Denoising diffusion implicit models},
  author={Song, Jiaming and Meng, Chenlin and Ermon, Stefano},
  journal={arXiv preprint arXiv:2010.02502},
  year={2020}
}

@inproceedings{rombach2022high,
  title={High-resolution image synthesis with latent diffusion models},
  author={Rombach, Robin and Blattmann, Andreas and Lorenz, Dominik and Esser, Patrick and Ommer, Bj{\"o}rn},
  booktitle=CVPR,
  pages={10684--10695},
  year={2022}
}

@inproceedings{zhang2023adding,
  title={Adding conditional control to text-to-image diffusion models},
  author={Zhang, Lvmin and Rao, Anyi and Agrawala, Maneesh},
  booktitle=ICCV,
  pages={3836--3847},
  year={2023}
}

@article{itoh1982analysis,
  title={Analysis of the phase unwrapping algorithm},
  author={Itoh, Kazuyoshi},
  journal=AO,
  volume={21},
  number={14},
  pages={2470--2470},
  year={1982}
}

@inproceedings{woo2018cbam,
  title={{CBAM}: Convolutional block attention module},
  author={Woo, Sanghyun and Park, Jongchan and Lee, Joon-Young and Kweon, In So},
  booktitle=ECCV,
  pages={3--19},
  year={2018}
}

@article{dosovitskiy2020image,
  title={An image is worth 16 $\times$ 16 words: Transformers for image recognition at scale},
  author={Dosovitskiy, Alexey and Beyer, Lucas and Kolesnikov, Alexander and Weissenborn, Dirk and Zhai, Xiaohua and Unterthiner, Thomas and Dehghani, Mostafa and Minderer, Matthias and Heigold, Georg and Gelly, Sylvain and others},
  journal={arXiv preprint arXiv:2010.11929},
  year={2020}
}

@inproceedings{he2016deep,
  title={Deep residual learning for image recognition},
  author={He, Kaiming and Zhang, Xiangyu and Ren, Shaoqing and Sun, Jian},
  booktitle=CVPR,
  pages={770--778},
  year={2016}
}

@article{ulyanov2016instance,
  title={Instance normalization: The missing ingredient for fast stylization},
  author={Ulyanov, Dmitry and Vedaldi, Andrea and Lempitsky, Victor},
  journal={arXiv preprint arXiv:1607.08022},
  year={2016}
}

@inproceedings{zhu2019deformable,
  title={Deformable {ConvNets} v2: More deformable, better results},
  author={Zhu, Xizhou and Hu, Han and Lin, Stephen and Dai, Jifeng},
  booktitle=CVPR,
  pages={9308--9316},
  year={2019}
}

@article{hinton2006reducing,
  title={Reducing the dimensionality of data with neural networks},
  author={Hinton, Geoffrey E and Salakhutdinov, Ruslan R},
  journal={Science},
  volume={313},
  number={5786},
  pages={504--507},
  year={2006}
}

@article{loshchilov2017decoupled,
  title={Decoupled weight decay regularization},
  author={Loshchilov, Ilya and Hutter, Frank},
  journal={arXiv preprint arXiv:1711.05101},
  year={2017}
}

@article{loshchilov2016sgdr,
  title={{SGDR}: Stochastic gradient descent with warm restarts},
  author={Loshchilov, Ilya and Hutter, Frank},
  journal={arXiv preprint arXiv:1608.03983},
  year={2016}
}

@article{mantiuk2023hdr,
  title={{HDR-VDP-3}: A multi-metric for predicting image differences, quality and contrast distortions in high dynamic range and regular content},
  author={Mantiuk, Rafal K and Hammou, Dounia and Hanji, Param},
  journal={arXiv preprint arXiv:2304.13625},
  year={2023}
}

@inproceedings{hanji2022comparison,
  title={Comparison of single image {HDR} reconstruction methods-the caveats of quality assessment},
  author={Hanji, Param and Mantiuk, Rafal and Eilertsen, Gabriel and Hajisharif, Saghi and Unger, Jonas},
  booktitle=SIGGRAPH,
  pages={1--8},
  year={2022}
}

@article{li2021benchmarking,
  title={Benchmarking detection transfer learning with vision {Transformers}},
  author={Li, Yanghao and Xie, Saining and Chen, Xinlei and Dollar, Piotr and He, Kaiming and Girshick, Ross},
  journal={arXiv preprint arXiv:2111.11429},
  year={2021}
}

@inproceedings{lindner2023lightweight,
  title={Lightweight video denoising using aggregated shifted window attention},
  author={Lindner, Lydia and Effland, Alexander and Ilic, Filip and Pock, Thomas and Kobler, Erich},
  booktitle=WACV,
  pages={351--360},
  year={2023}
}
}

\vspace{-10mm}

\begin{IEEEbiography}
[{\includegraphics[width=1in,height=1.25in,clip,keepaspectratio]{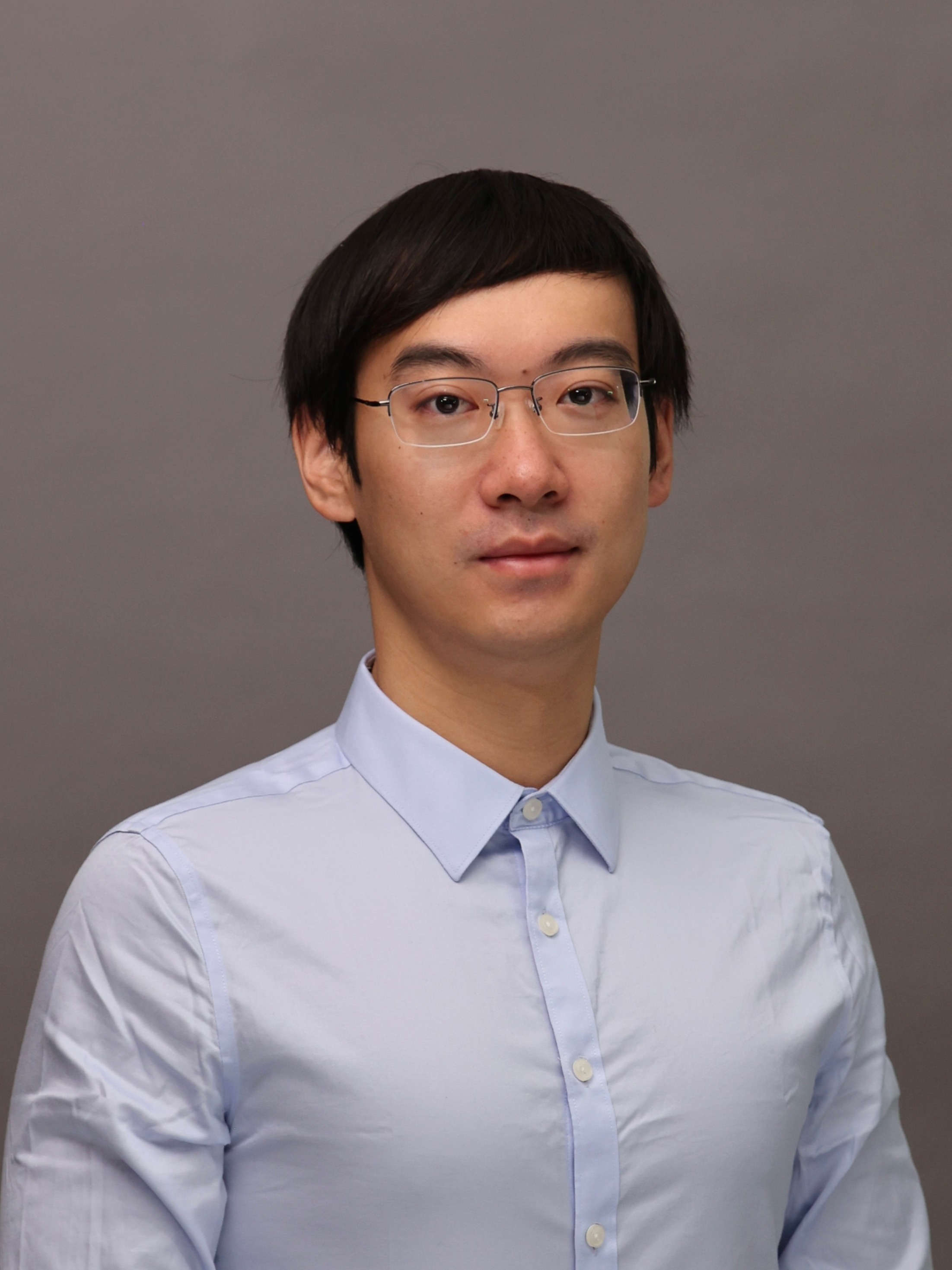}}]
{Chu Zhou} received the B.E. degree from Huazhong University of Science and Technology in 2019 and the Ph.D. degree from School of Intelligence Science and Technology, Peking University in 2024. He is currently an assistant professor by special appointment at Digital Contents and Media Sciences Research Division, National Institute of Informatics. His research interest includes computational photography and computer vision, with a focus on unconventional camera-based vision.
\end{IEEEbiography}
\vspace{-10mm}

\begin{IEEEbiography}
[{\includegraphics[width=1in,height=1.25in,clip,keepaspectratio]{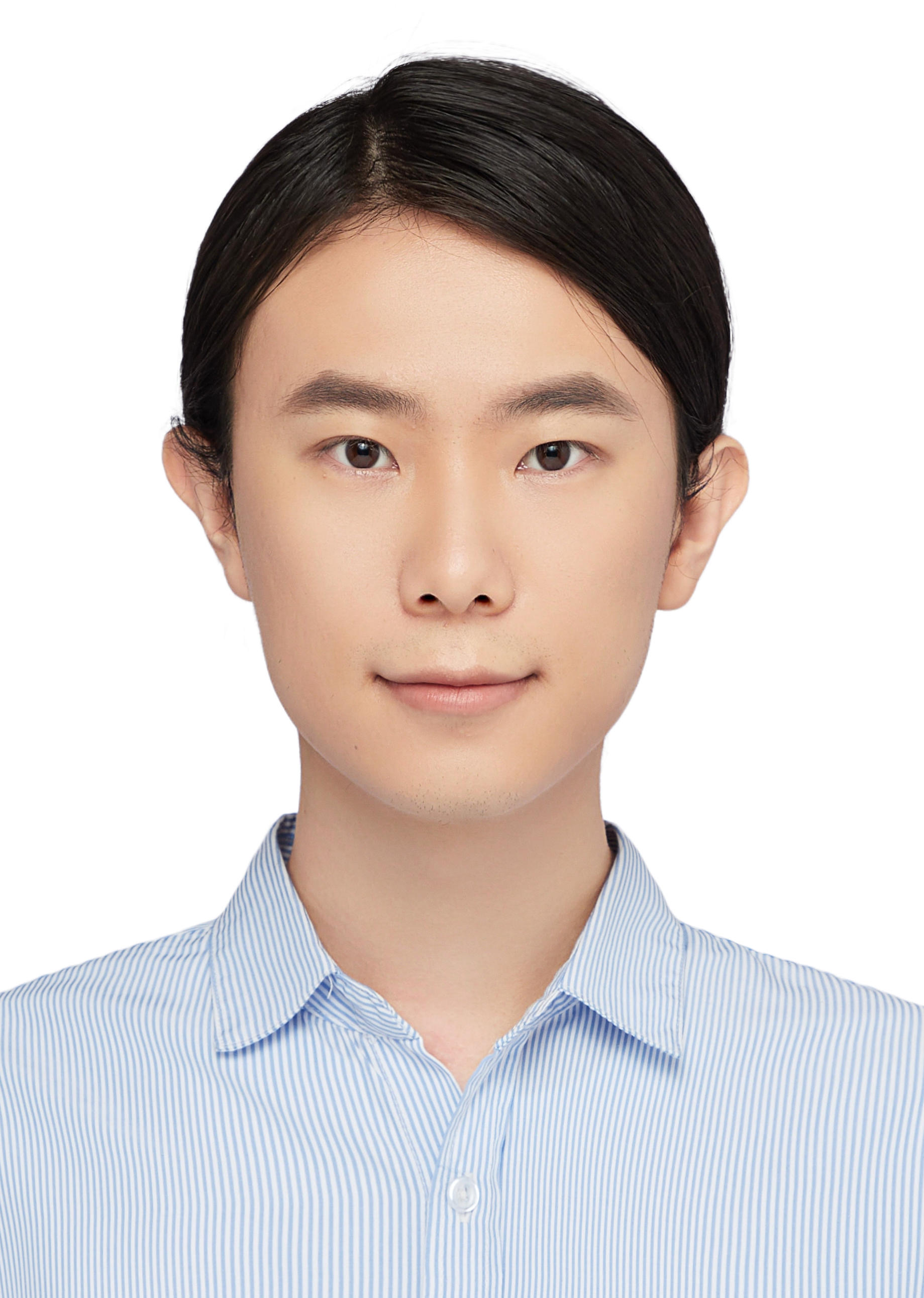}}]
{Siqi Yang} received the B.S. degree in computer science from the Turing Honor Class, School of Electronics Engineering and Computer Science, Peking University, Beijing, China, in 2022. He is currently pursuing the Ph.D. degree with the Institute for Artificial Intelligence, Peking University, Beijing, under the supervision of Prof. Zhaofei Yu and Prof. Boxin Shi. His current research interests include computational photography, neuromorphic imaging, inverse rendering, and video generation.

\end{IEEEbiography}
\vspace{-10mm}

\begin{IEEEbiography}
[{\includegraphics[width=1in,height=1.25in,clip,keepaspectratio]{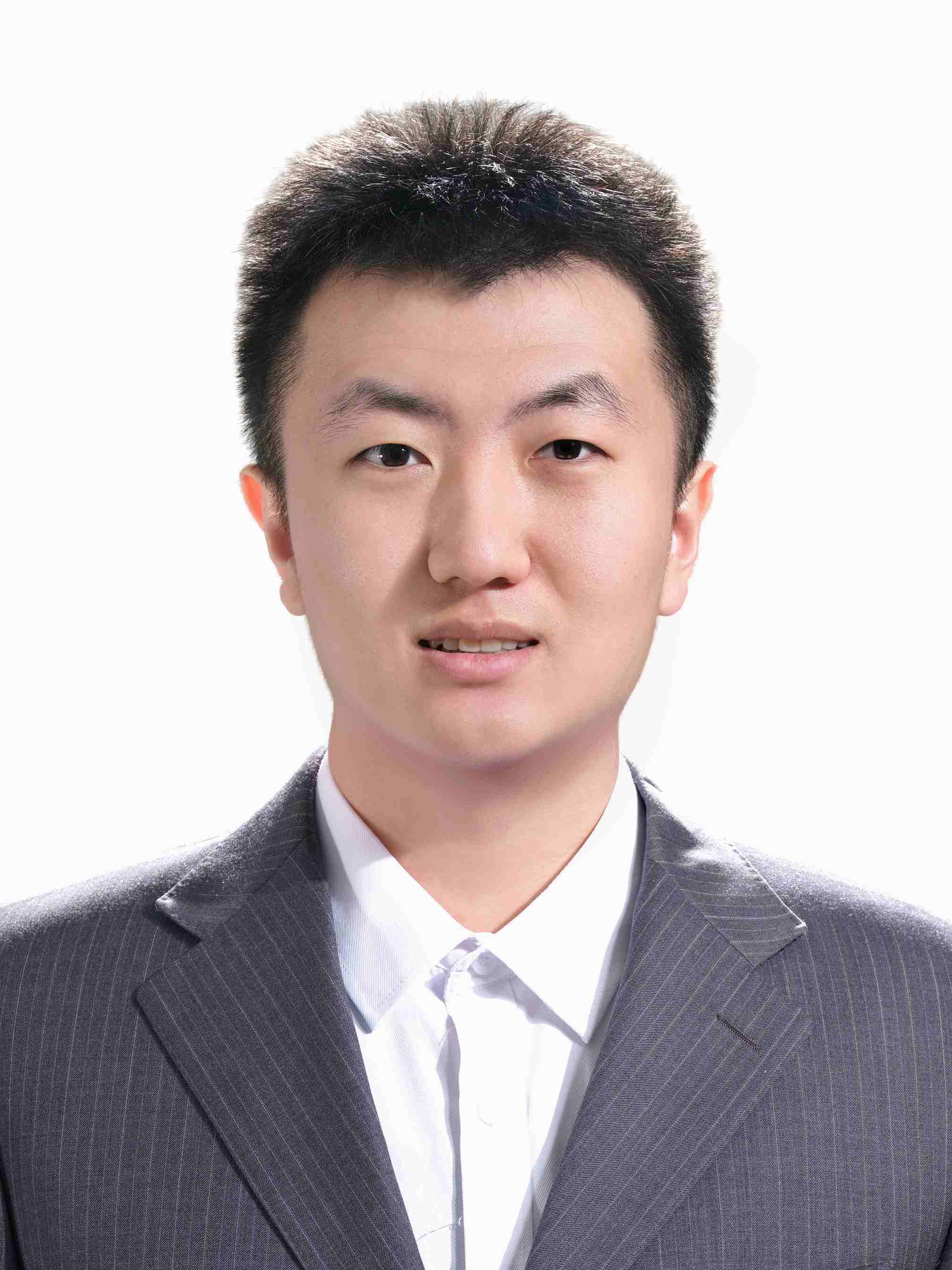}}]
{Kailong Zhang} received his B.S. degree from Beijing University of Posts and Telecommunications in 2024. He is currently working toward his master's degree at Beijing University of Posts and Telecommunications. His research interests are centered around computational photography and image generation.
\end{IEEEbiography}
\vspace{-10mm}

\begin{IEEEbiography}
[{\includegraphics[width=1in,height=1.25in,clip,keepaspectratio]{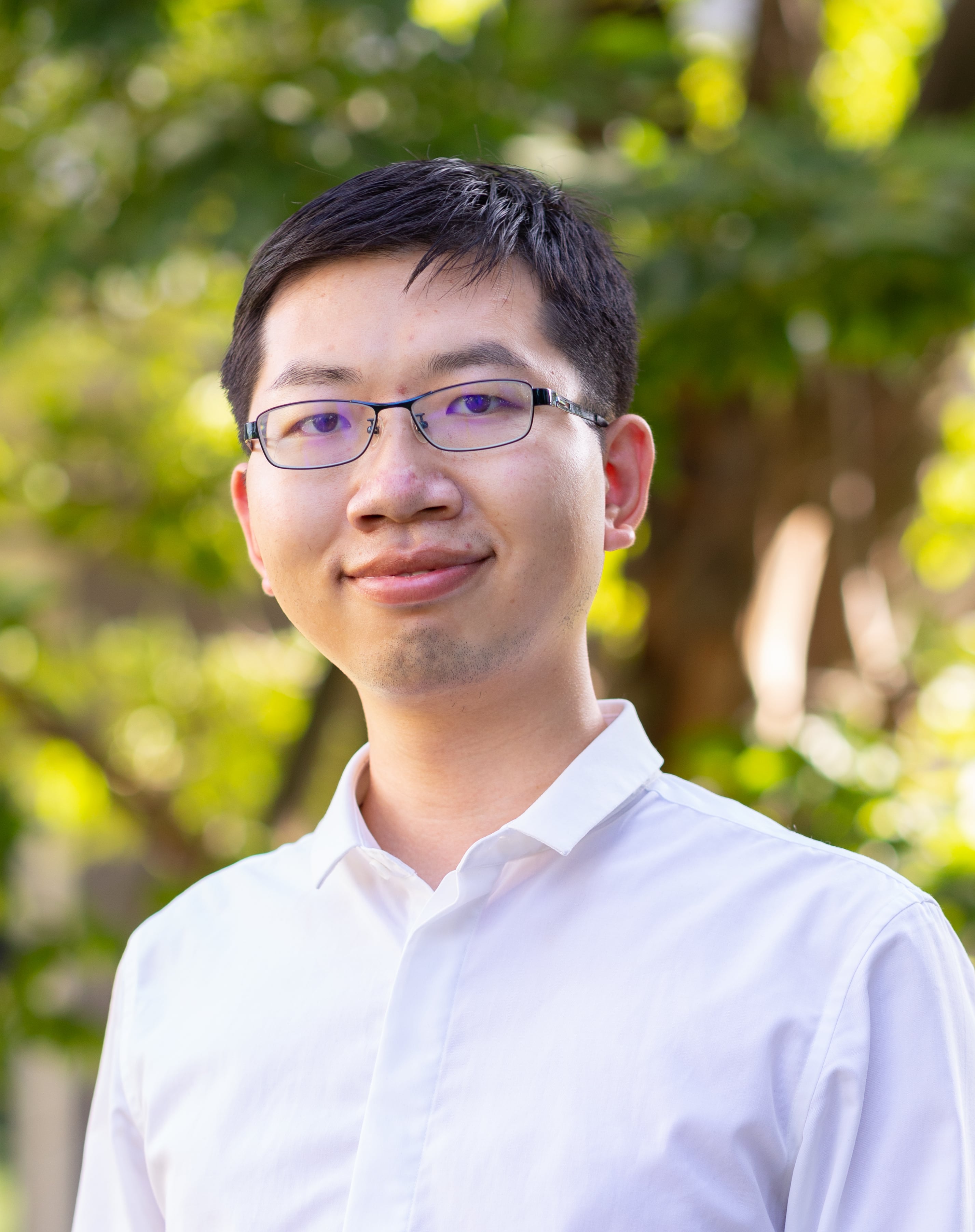}}]
{Heng Guo} received the BE and ME degrees from University of Electronic Science and Technology, and the PhD degree from Osaka University, in 2015, 2018, and 2022. He is currently a specially-appointed research Professor at Beijing University of Posts and Telecommunications (BUPT). Before joining BUPT, he was a specially-appointed assistant professor at Osaka University from 2022 to 2023. His research interest includes computational photography, physics-based computer vision, and 3D reconstruction. He served as reviewers of CVPR, ICCV, ECCV, TPAMI, IJCV.
\end{IEEEbiography}
\vspace{-10mm}

\begin{IEEEbiography}
[{\includegraphics[width=1in,height=1.25in,clip,keepaspectratio]{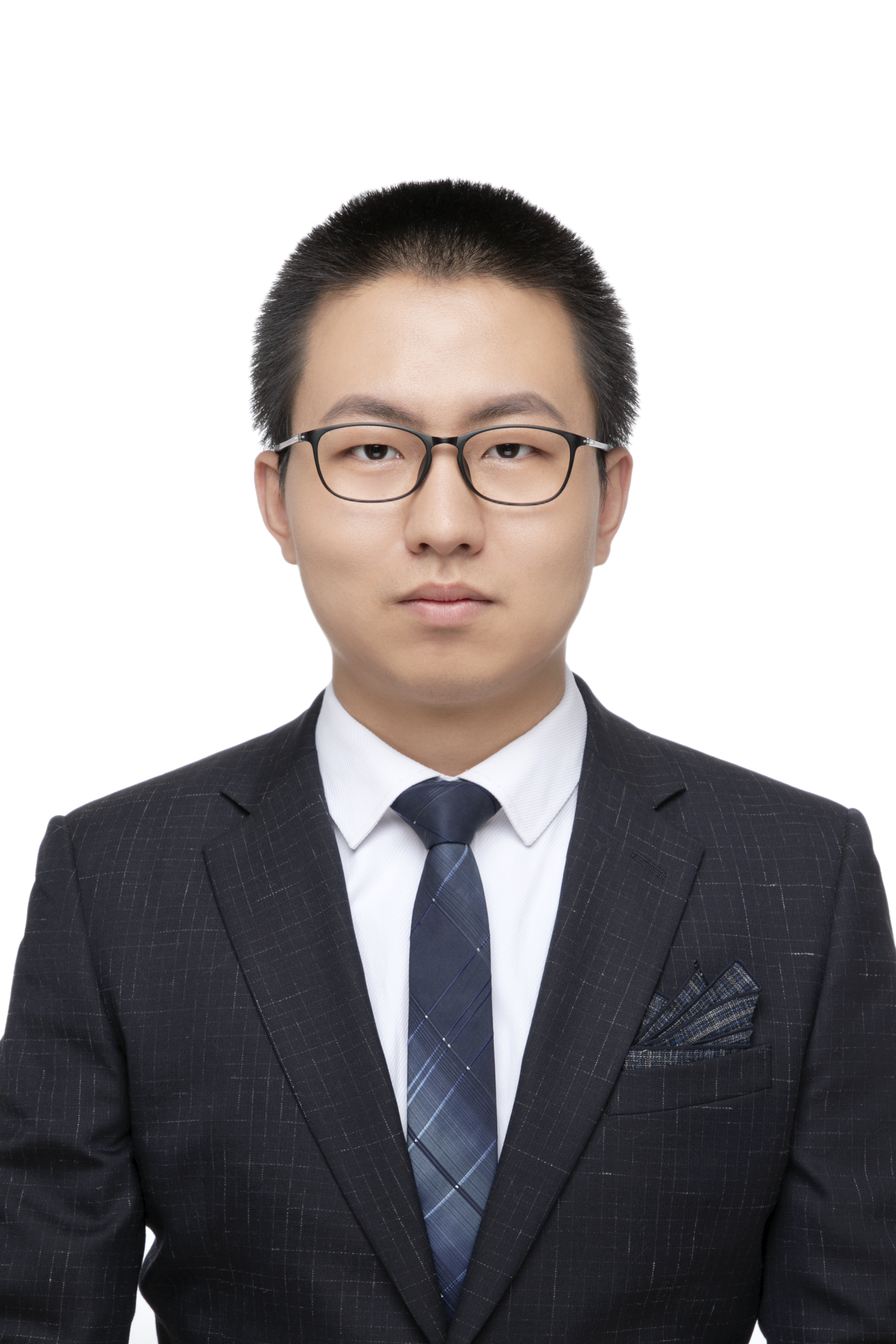}}]
{Zhaofei Yu} (Member, IEEE) received the B.S. degree from the Hong Shen Honors School, College of Optoelectronic Engineering, Chongqing University, Chongqing, China, in 2012, and the Ph.D. degree from the Automation Department, Tsinghua University, Beijing, China, in 2017. He is currently an Assistant Professor with the Institute for Artificial Intelligence, Peking University, Beijing. His current research interests include artificial intelligence, brain-inspired computing, and computational neuroscience.

\end{IEEEbiography}
\vspace{-10mm}

\begin{IEEEbiography}
[{\includegraphics[width=1in,height=1.25in,clip,keepaspectratio]{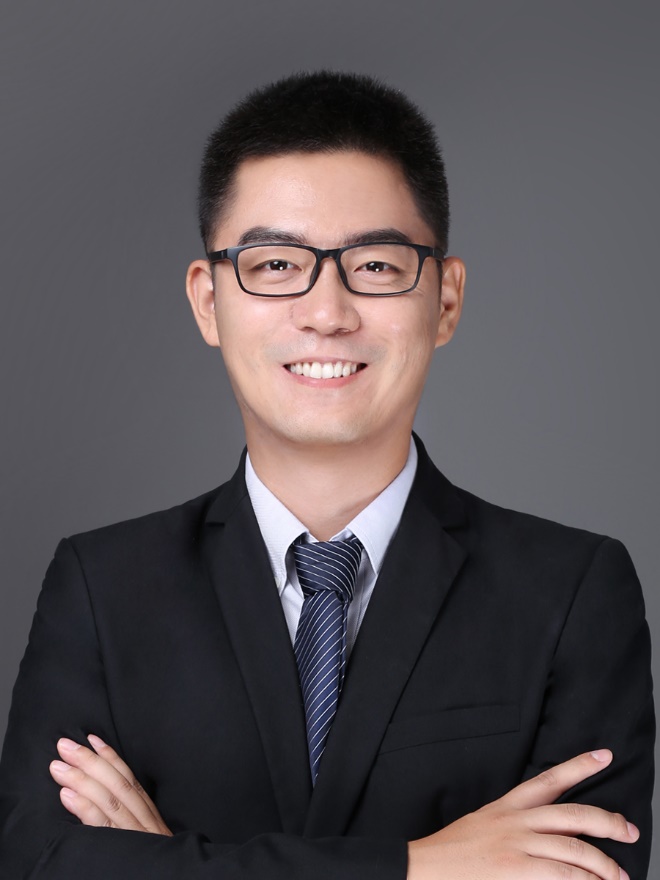}}]
{Boxin Shi} received the B.E. degree from the Beijing University of Posts and Telecommunications, the M.E. degree from Peking University, and the Ph.D. degree from the University of Tokyo, in 2007, 2010, and 2013. He is currently a Boya Young Fellow Associate Professor (with tenure) and Research Professor at Peking University, where he leads the Camera Intelligence Lab. Before joining PKU, he did research with MIT Media Lab, Singapore University of Technology and Design, Nanyang Technological University, National Institute of Advanced Industrial Science and Technology, from 2013 to 2017. His papers were awarded as Best Paper, Runners-Up at CVPR 2024, ICCP 2015, and selected as Best Paper candidate at ICCV 2015. He is an associate editor of TPAMI/IJCV and an area chair of CVPR/ICCV/ECCV. He is a senior member of IEEE.
\end{IEEEbiography}
\vspace{-10mm}

\begin{IEEEbiography}
[{\includegraphics[width=1in,height=1.25in,clip,keepaspectratio]{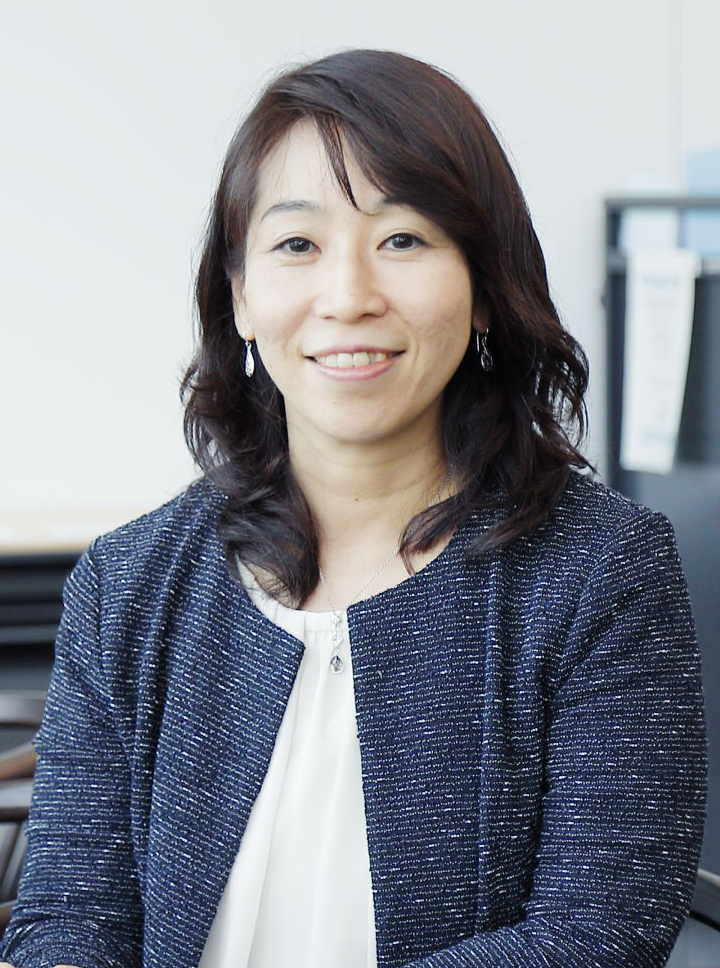}}]
{Imari Sato} received the B.S. degree in policy management from Keio University, Tokyo, Japan, in 1994, and the M.S. and Ph.D. degrees in interdisciplinary Information Studies from the University of Tokyo, Tokyo, in 2002 and 2005, respectively. She was a visiting scholar with the Robotics Institute of Carnegie Mellon University, Pittsburgh, PA, USA. In 2005, she joined the National Institute of Informatics, where she is currently a professor/director of the Digital Contents and Media Sciences Research Division. She is a professor at the University of Tokyo and a visiting professor at the Tokyo Institute of Technology, Tokyo, Japan. Her primary research interests include physics-based vision, spectral analysis, image-based modeling, and medical image analysis. She received various research awards, including the Young Scientists’ Prize from the Commendation for Science and Technology by the Minister of Education, Culture, Sports, Science and Technology in 2009, and the Microsoft Research Japan New Faculty Award in 2011.
\end{IEEEbiography}

\end{document}